\PassOptionsToPackage{table}{xcolor}
\documentclass[10pt]{article} 
\usepackage[accepted]{tmlr}

\usepackage{amsmath,amsfonts,bm}

\def\eqref#1{equation~\ref{#1}}

\def\1{\bm{1}}

\DeclareMathAlphabet{\mathsfit}{\encodingdefault}{\sfdefault}{m}{sl}
\SetMathAlphabet{\mathsfit}{bold}{\encodingdefault}{\sfdefault}{bx}{n}

\usepackage[utf8]{inputenc} 
\usepackage[T1]{fontenc}    
\usepackage{hyperref}       
\usepackage{url}            
\usepackage{booktabs}       
\usepackage{amsfonts}       
\usepackage{nicefrac}       
\usepackage{microtype}      

\usepackage{array}
\usepackage{xcolor}         
\usepackage{multirow}
\usepackage{bm}
\usepackage{arydshln}
\usepackage{graphicx}
\usepackage{caption}
\usepackage{caption}
\usepackage{tcolorbox}
\usepackage{etoolbox}
\usepackage{hhline} 
\usepackage{pdfpages}

\usepackage[T1]{fontenc}    
\usepackage{url}
\usepackage{adjustbox}
\usepackage{array}
\usepackage{pifont}
\usepackage{booktabs}    
\usepackage{multirow}    
\usepackage{rotating}    
\usepackage{subcaption}  
\usepackage{graphicx}    
\usepackage{amsmath}     
\usepackage{amssymb}     
\usepackage{wrapfig}
\usepackage{enumitem}
\usepackage{float}
\usepackage{placeins}

\usepackage{multicol}        
\usepackage{enumitem}        

\usepackage{soul}   

\newif\ifshowrevisionhighlight
\showrevisionhighlightfalse
\ifshowrevisionhighlight
  \definecolor{revblue}{HTML}{DDF7FF}
\else
  \definecolor{revblue}{HTML}{FFFFFF}
\fi
\sethlcolor{revblue}
\soulregister\citep7
\soulregister\citet7
\soulregister\cite7
\soulregister\ref7
\soulregister\url7
\soulregister\texttt7
\soulregister\textbf7
\soulregister\textit7
\soulregister\shortstack7
\ifshowrevisionhighlight
  \DeclareRobustCommand{\rev}[1]{\begingroup\sethlcolor{revblue}\hl{#1}\endgroup}
\else
  \DeclareRobustCommand{\rev}[1]{#1}
\fi

\newcommand{\topic}[1]{\noindent\textbf{#1}\quad}

\title{FreeFuse: Multi-Subject LoRA Fusion via Adaptive Token-Level Routing at Test Time}

\author{\name Yaoli Liu \email yaoliliu8@gmail.com \\
      \addr State Key Laboratory of CAD\&CG\\
      Zhejiang University
      \AND
      \name Yao-Xiang Ding\thanks{Corresponding author.} \email dingyx.gm@gmail.com \\
      \addr State Key Laboratory of CAD\&CG\\
      Zhejiang University
      \AND
      \name Kun Zhou \email kunzhou@acm.org\\
      \addr State Key Laboratory of CAD\&CG\\
      Zhejiang University
      }

\def\month{09}  
\def\year{2026} 
\def\openreview{\url{https://openreview.net/forum?id=Bvela8OAMc}} 

\begin{document}

\maketitle

\begin{abstract}

This paper proposes FreeFuse, a training-free framework for multi-subject text-to-image generation through automatic fusion of multiple subject LoRAs. \rev{In contrast to prior studies that focus on retraining LoRAs to alleviate feature conflicts, our analysis shows that spatially routing LoRA residuals to their intended semantic regions provides an effective mechanism for suppressing direct cross-region LoRA interference while preserving the base model's global contextual reasoning.} Accordingly, we implement Adaptive Token-Level Routing during the inference phase. However, obtaining reliable routing regions remains challenging. Existing methods that rely on text-image latent association, such as raw cross-attention or concept-level similarity matching, often suffer from sparse activations, hole artifacts, and unstable localization when handling visually similar subjects, leading to incomplete or ambiguous subject masks. To address these issues, we introduce FreeFuseAttn, a mechanism that exploits the flow matching model's intrinsic semantic alignment to dynamically match subject-specific tokens to their corresponding spatial regions at early denoising timesteps, thereby bypassing the need for external segmentors. FreeFuse distinguishes itself through high practicality: it necessitates no additional training, model modifications, or user-defined spatial constraints. Users need only provide subject activation words to achieve seamless integration into standard workflows. Extensive experiments validate that FreeFuse outperforms existing approaches in both identity preservation and compositional fidelity. The project page is at: \url{https://future-item.github.io/FreeFuse/}.
\end{abstract}

  \begin{figure*}[t]
  \centering
  \centering
  \includegraphics[width=1\textwidth]{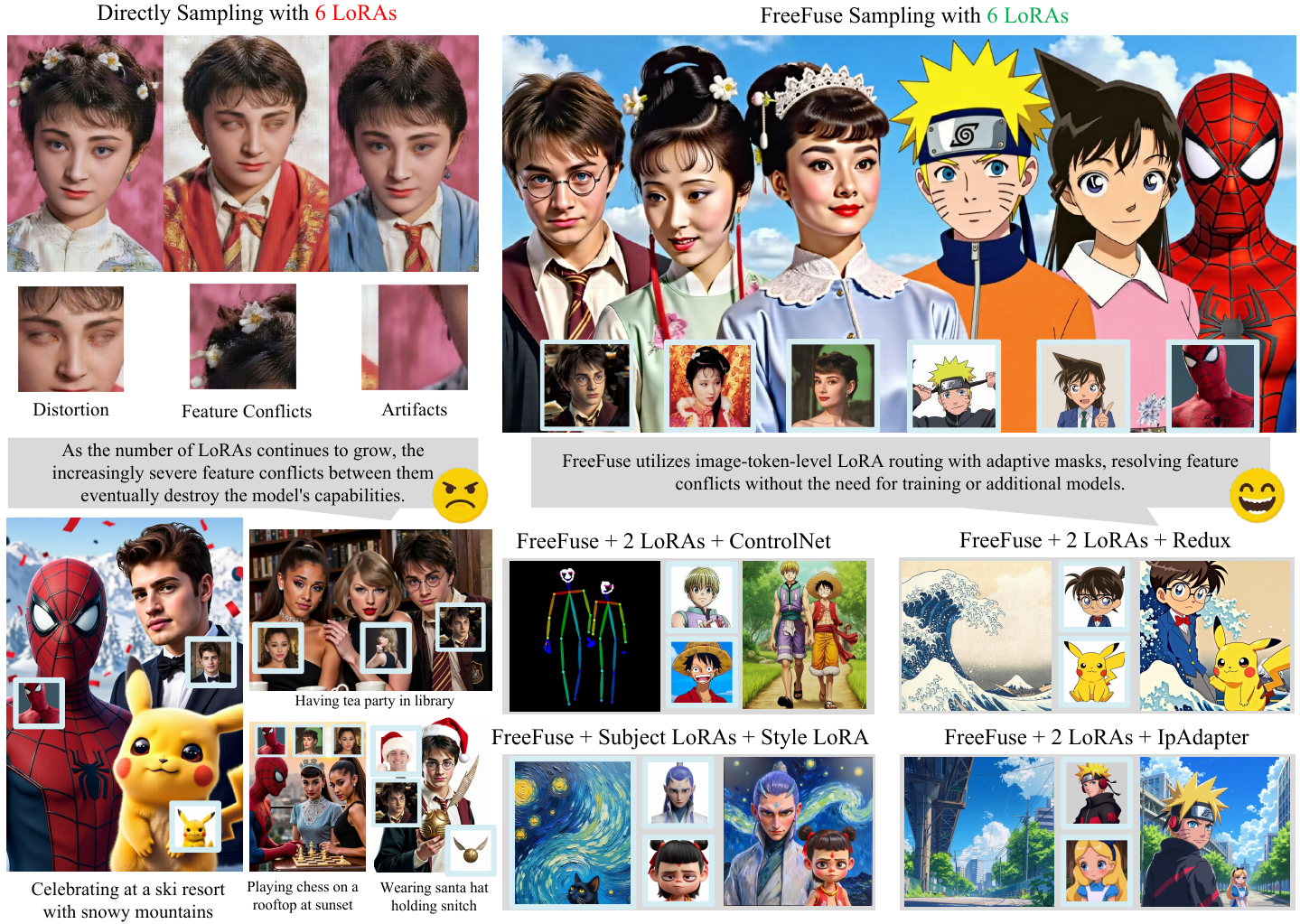}
  \captionsetup{width=1\linewidth}
  \caption{\textbf{Multi-subject-LoRA generation with FreeFuse}. We propose a robust, training-free framework that mitigates feature conflicts among multiple LoRAs by spatially constraining their influence. Our core mechanism, FreeFuseAttn, fuses semantic-driven cross-attention with cohesion-driven token similarity to generate contiguous subject masks, effectively reducing "hole" artifacts. The framework achieves high identity fidelity and is plug-and-play compatible with mainstream control modules without requiring auxiliary networks.}
  \label{fig:teaser}
\end{figure*}

\section{Introduction}
\label{sec:intro}

Large-scale text-to-image (T2I) models such as 
FLUX.1-dev~\citep{flux2024,cai2025hidream} have demonstrated remarkable performance in general T2I tasks. To enhance their capability for personalized generation, Low-Rank Adaptation (LoRA)~\citep{hu2022lora} has emerged as a preferred approach due to its precise fine-tuning quality and computational efficiency in both training and inference. LoRA also enables a simple way for multi-subject generation: Multiple subject LoRAs can be directly combined on the pretrained T2I models for multi-subject generation. 
However, this straightforward approach can lead to performance degradation, with the appearance of feature conflicts and deterioration~\citep{shah2024ziplora,kong2024omg,kwon2024concept,meral2024clora,dalva2025lorashop,po2024orthogonaladaptationmodularcustomization}, making multi-subject LoRA fusion a challenging problem. 

Prior works on multi-LoRA generation~\citep{shah2024ziplora,gu2023mix,kong2024omg,meral2024clora,kwon2024concept} rely on designated techniques such as retraining, additional trainable parameters, external segmentation models or requiring users to provide template prompts or directly constrain the regions where LoRAs take effect, yet still struggle with multi-subject generation in complex scene. While recent approaches~\citep{kong2024omg, kwon2024concept, dalva2025lorashop, meral2024clora} mitigate conflicts via heavy inference interventions like noise or residual blending, these methods often compromise global coherence and inflate latency. We identify the root cause of feature conflict not as a generation artifact, but as the indiscriminate broadcasting of LoRA parameter updates ($\Delta\theta$). \rev{We demonstrate that routing additive LoRA residuals to their intended semantic regions effectively suppresses direct cross-region interference.}

FreeFuse operates in two phases. In the first phase, corroborated by prior findings~\citep{kwon2024concept, helbling2025conceptattentiondiffusiontransformerslearn,zhang2024spdiffusion, choi2022perception} and our empirical observations, we introduce FreeFuseAttn, which capitalizes on the base model's latent segmentation capacity to construct an Image-Token-Level Router strictly during the early denoising stages. Simultaneously, we derive an Attention Bias from these high-fidelity segmentation cues to encourage robust alignment between spatial regions and subject semantics. \rev{In the second phase, the router orchestrates the multi-LoRA inference process by assigning each spatial token to a semantic subject/group and applying the adapters associated with that group. This supports both one-LoRA-per-subject and multi-adapter descriptions of a single subject, while still enforcing exclusivity between competing subject groups.} Crucially, under the guidance of the Attention Bias, the model prioritizes subject-specific semantics, thereby significantly reducing concept bleeding.

In summary, our core contributions to the community include:
(1) \rev{We provide a mechanism analysis demonstrating that spatially constraining additive LoRA residuals to target regions is an effective and empirically supported way to mitigate feature conflicts among multiple subjects.}
(2) FreeFuseAttn, which mitigates the `hole' artifacts in latent space segmentation by fusing semantic-driven cross-attention with cohesion-driven token similarity. This approach ensures more complete and contiguous subject masks, outperforming cross-attention, ConceptAttention or SP-Attn methods in multi-object generation scenarios.
(3) A robust, training-free multi-subject generation framework that fully harnesses the intrinsic capabilities of the base model, eliminating the need for external segmentation modules or auxiliary networks. Due to its lightweight design, our approach offers seamless plug-and-play compatibility with mainstream control modules, such as spatial guidance tools (e.g., ControlNet), reference adapters (e.g., IP-Adapter, Redux), and aesthetic fine-tuners (e.g., Style LoRAs).
(4) Extensive experiments confirm FreeFuse's superiority. Quantitative results establish a new state-of-the-art in identity preservation and compositional fidelity without compromising global image aesthetics. 

\section{Related Work}
\label{sec:rw}

\subsection{Text-to-Image Diffusion Model}
In recent years, image generation models have advanced rapidly, evolving from early GAN-based models~\citep{goodfellow2014generative,arjovsky2017wasserstein,karras2019style,karras2020analyzing} to U-Net-based diffusion models~\citep{ronneberger2015u,ho2020denoising,song2020denoising,rombach2022high}, and further to the widely adopted DiT-based diffusion models~\citep{podell2023sdxl,peebles2023scalable,esser2024scaling,flux2024,cai2025hidream,wu2025qwen,labs2025flux1kontextflowmatching}. With the continuous growth of model size, training scale, and architectural improvements, large-scale DiT-based models such as 
FLUX.1-dev~\citep{flux2024} have become leaders among open-source models, while also driving research into customized generation, local editing, and style transfer.

\subsection{Personalized Image Generation}
Customized generation in diffusion models has been extensively studied. Textual inversion~\citep{gal2022image} methods encode rich semantic information into one or several text tokens through training. IP-Adapter~\citep{ye2023ip},FLUX-Redux~\citep{flux2024} and InstantID~\citep{wang2024instantid} instead train a generalizable module that directly takes one or more images and encodes their semantics into features aligned with the text or latent space. DreamBooth~\citep{ruiz2023dreambooth} introduces new concepts by fine-tuning diffusion network weights. With the wide adoption of LoRA~\citep{hu2022lora} as an efficient fine-tuning method, fine-tuning open-source diffusion models with LoRA for customized generation has become a common choice among community users. Numerous works further improve LoRA or its training strategies, such as LyCORIS~\citep{yeh2023navigating}, QLoRA~\citep{dettmers2023qlora}, ED-LoRA~\citep{gu2023mix}, and SD-LoRA~\citep{wu2025sd}, but LoRA itself remains the most widely used solution.

\subsection{Multi Concept Generation}

Early attempts such as ZipLoRA~\citep{shah2024ziplora} and K-LoRA~\citep{ouyang2025k} fuse multiple LoRAs prior to inference. While successful in style transfer, these methods exhibit limited performance in preserving identities during multi-subject generation. Multi-LoRA~\citep{zhong2024multi} introduces switch and composite strategies to mitigate conflicts; however, it struggles to delineate boundaries in complex multi-character scenarios. To enforce spatial isolation, methods like OMG~\citep{kong2024omg}, Concept Weaver~\citep{kwon2024concept}, and FlipConcept~\citep{woo2025flipconcept} employ auxiliary segmentation models combined with noise blending, yet these approaches often falter when delineating visually similar subjects (e.g., two men). While TokenVerse~\citep{garibi2025tokenverse} attempts to fuse concepts via Token Modulation, it remains suboptimal when handling semantic symmetry. Conversely, Mix-of-Show~\citep{gu2023mix}, Orthogonal Adaptation~\citep{po2024orthogonaladaptationmodularcustomization}, and LoRACLR~\citep{simsar2025loraclr} require retraining LoRAs with manual spatial specifications; although they achieve high identity preservation, their rigid spatial constraints severely compromise compositional flexibility. Similarly, DreamRelation~\citep{shi2025dreamrelationbridgingcustomizationrelation} offers fine-grained control but imposes a significant manual burden on the user. To circumvent these retraining costs and manual interventions, recent training-free approaches, including CLoRA~\citep{meral2024clora}, $MC^2$~\citep{jiang2025mc}, and LoRAShop~\citep{dalva2025lorashop}, leverage cross-attention maps to derive concept masks. However, their performance degrades in complex scenes due to attentional leakage or maps deviating from semantic expectations. Furthermore, unified understanding-generation models like OmniGen~\citep{xiao2025omnigen}, Xverse~\citep{chen2025xverse}, UNO~\citep{wu2025uno}, and UMO~\citep{cheng2025umo} explore multi-concept generation but are often restricted to a single reference image for each subject, leading to the loss of fine-grained identity details or layout overfitting. Distinct from these approaches, FreeFuse is positioned as a fully training-free and auxiliary-free LoRA fusion framework. By maximizing the model's intrinsic cross-modal alignment within the latent space, we synthesize high-quality multi-subject images with flexible, natural layouts, effectively liberating users from the burdens of LoRA retraining and manual spatial specification.

\section{Methodology}
\begin{figure*}[t]
  \centering
  \centering
  \includegraphics[width=1\textwidth]{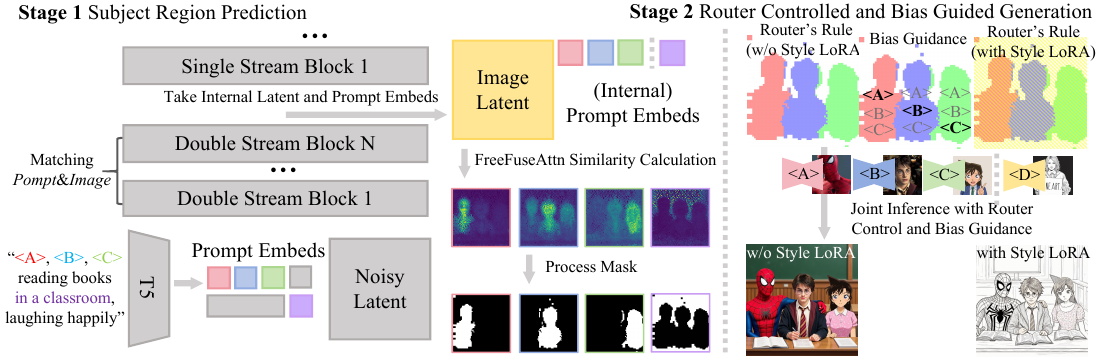}
  \captionsetup{width=1\linewidth}
  \caption{\textbf{The FreeFuse Pipeline.} In \textbf{Phase 1}, we employ FreeFuseAttn to extract robust subject masks, which are subsequently processed into a spatial Router and Attention Bias. In \textbf{Phase 2}, \rev{the Router enforces subject LoRA exclusivity per token to suppress direct cross-region LoRA interference}, while the Bias mechanism actively reduces concept bleeding by ensuring precise semantic-spatial alignment.}
  \label{fig:method}
\end{figure*}

In this section, we present FreeFuse, a comprehensive framework designed to resolve identity conflicts in multi-subject generation. We begin by demonstrating that naively aggregating LoRA outputs induces severe feature interference, and pointing out that \rev{spatially routing additive LoRA residuals to their semantic regions directly suppresses the major conflict path.} Building on this insight, we systematically explore the intrinsic segmentation capabilities of Flow Matching models to localize subjects without external supervision, proposing \textbf{FreeFuseAttn} as a robust attention-based localization mechanism. Finally, we integrate these components into a two-stage pipeline: primarily, we leverage the model's intrinsic capabilities to predict precise subject masks; subsequently, we deploy a token-level Router and an attention Bias mechanism to strictly enforce local LoRA activation and suppress concept bleeding, enabling high-fidelity multi-subject synthesis.

\subsection{Masking LoRA Outputs for Effective Subject Feature Preservation}
\label{sec:method_theory}

\begin{figure*}[t]
  \centering
  \begin{subfigure}{0.43\textwidth}
    \centering
    \includegraphics[width=\linewidth]{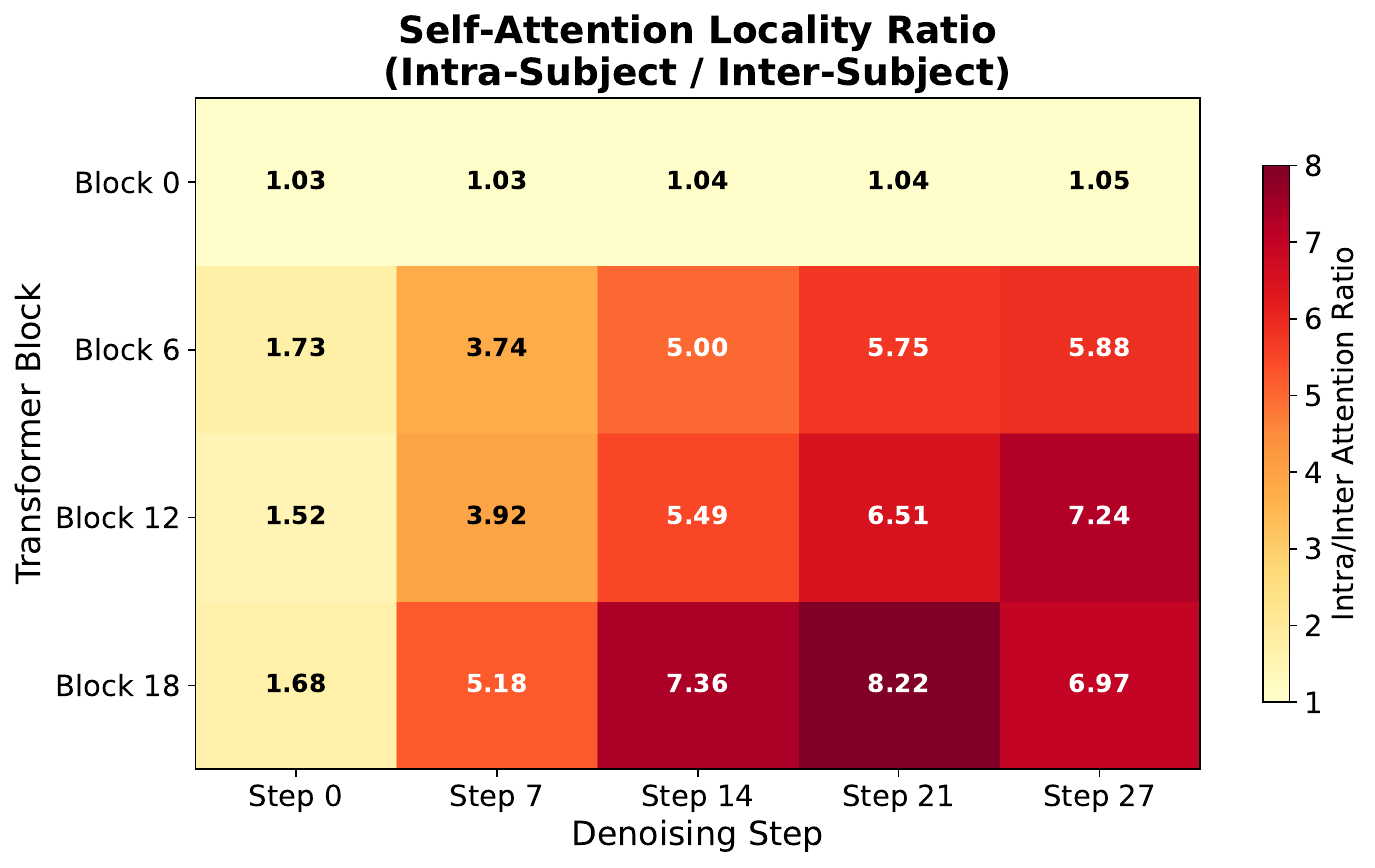}
  \end{subfigure}
  \hfill 
  \begin{subfigure}{0.56\textwidth}
    \centering
    \includegraphics[width=\linewidth]{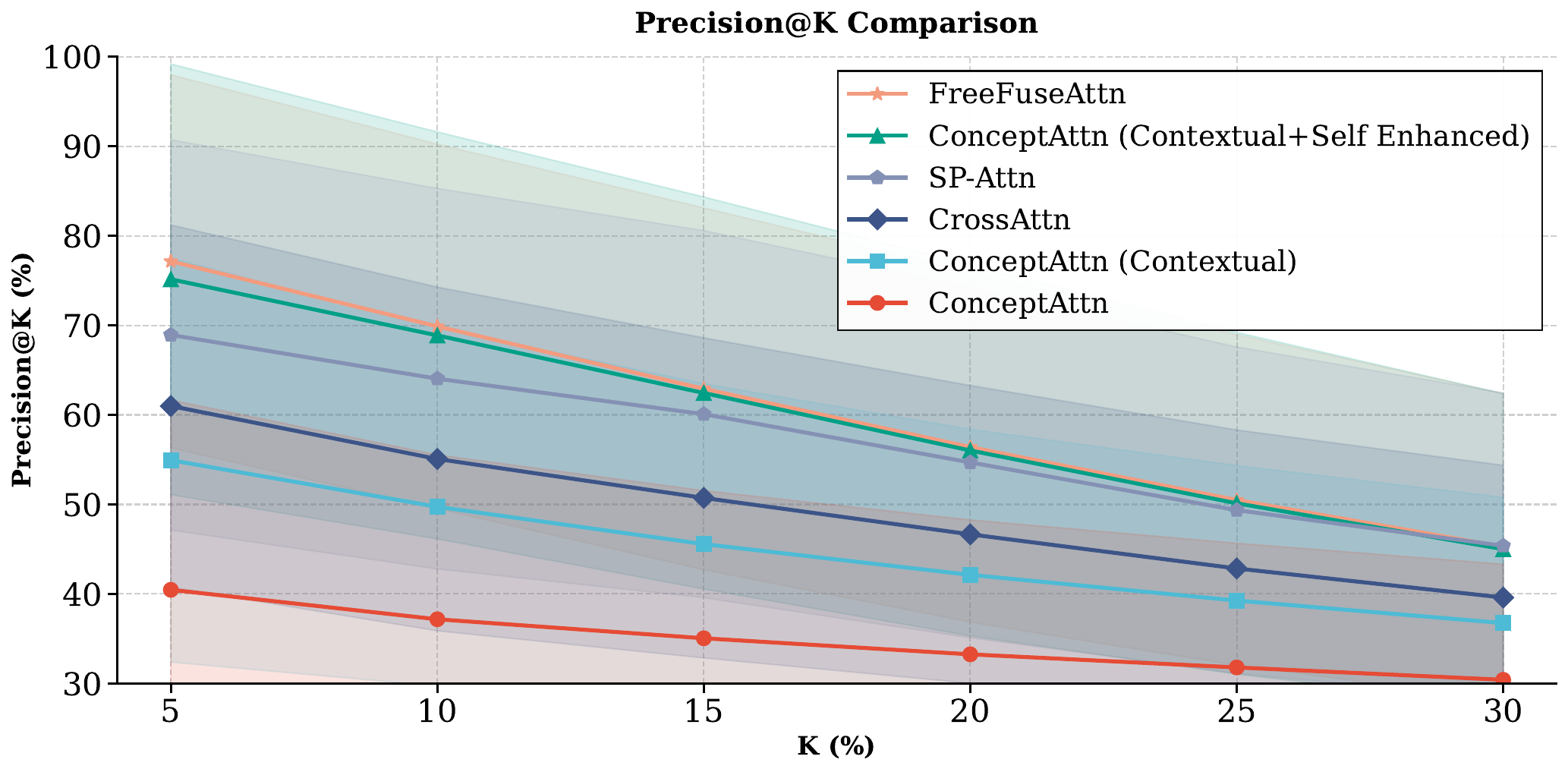}
  \end{subfigure}
  
  \captionsetup{width=1\linewidth} 
  \caption{\textbf{Left:} Empirical analysis of attention locality on 100 samples using SAM3 masks. The heatmap reports the ratio of intra- to inter-subject attention weights. While early blocks aggregate global context (ratio $\approx 1$), deep layers exhibit strong spatial isolation (ratio $\gg 1$), validating our LoRA masking strategy. \textbf{Right:} We evaluate the spatial alignment of different mechanisms against SAM3-generated ground truth masks over 300 samples. \textbf{Precision@K} measures the percentage of the top $K\%$ activated tokens that correctly fall within the subject's region. FreeFuseAttn demonstrates superior localization accuracy.}
  \label{fig:locality_and_precision_curve}
\end{figure*}

Consider \rev{$S$ semantic subject groups $\{G_1, \dots, G_S\}$, corresponding to spatial regions $\{R_1, \dots, R_S\}$ in the latent space. Each group $G_s$ can be associated with one or more LoRA adapters $\mathcal{A}_s$. This group-level formulation supports both the common one-subject-one-LoRA case and cases where a single subject is jointly described by multiple adapters, such as an identity LoRA together with an attribute LoRA or a style LoRA.} Directly merging all the LoRAs involves a naive summation of all LoRA outputs. This can introduce severe feature interference, as a token at position $p \in R_k$ is simultaneously perturbed by conflicting outputs from unrelated subject groups.

To enforce subject disentanglement, we utilize a spatial masking strategy. \rev{For a spatial token $p$, let $x_p$ denote the input hidden representation to the LoRA-augmented linear layer, $h_p$ denote the corresponding base-model output without LoRA residuals, and $h'_p$ denote the routed output after adding selected LoRA residuals. For any token position $p$, we retain only the contributions from adapters assigned to the corresponding semantic group:}
\begin{equation}
\colorbox{revblue}{$\displaystyle
\begin{aligned}
    h'_p = h_p + \sum_{s=1}^{S} \mathbb{I}(p \in R_s)
    \sum_{a \in \mathcal{A}_s} \Delta\theta_a(x_p),
\end{aligned}
$}
\end{equation}
where \rev{$\mathbb{I}(\cdot)$ is the indicator function and $\Delta\theta_a(x_p)$ denotes the additive residual produced by adapter $a$. This ensures that the LoRA feature update at region $R_k$ is governed only by the adapters assigned to group $G_k$, while LoRAs from competing groups are suppressed.}

A potential concern is that LoRA features from disjoint regions $R_{j \neq k}$ might still propagate into $R_k$ through the global aggregation of the self-attention mechanism:
\rev{For a self-attention layer with token features $X$, let $Q=XW_Q$, $K=XW_K$, and $V=XW_V$ denote the query, key, and value matrices. The attention matrix is $A=\mathrm{Softmax}(QK^\top/\sqrt{d})$, where $A_{p,q}$ is the attention weight from query token $p$ to value token $q$:}
\begin{equation}
    \text{Attn}(Q, K, V)_p = \sum_{q} A_{p,q} V_q,
\end{equation}
\rev{and $V_q$ is the value vector at token $q$. We do not claim that the base attention path prevents all cross-region information propagation; this global context exchange is useful for scene coherence. Instead, FreeFuse blocks the direct injection of subject-specific LoRA residuals into unrelated regions.} For the remaining indirect propagation through base attention, we leverage the intrinsic \textbf{\textit{spatial locality}} widely observed~\citep{raghu2022visiontransformerslikeconvolutional,caron2021emergingpropertiesselfsupervisedvision,helbling2025conceptattentiondiffusiontransformerslearn, tian2024diffuseattend} in DiT models. As visualized in Fig.~\ref{fig:locality_and_precision_curve} Left, while the initial layers (e.g., Block 0) exhibit a ratio near 1.0, indicating a global context aggregation necessary for layout initialization, the ratio rises sharply in deeper semantic layers (reaching $>7.0$ in Block 12-18). Crucially, this \textit{diagonal dominance} becomes most pronounced during the semantic-forming denoising steps, where the attention map $A$ exhibits a strong diagonal dominance thus tokens in region $R_k$ predominantly attend to other tokens within the same region:
\begin{equation}
    \sum_{q \in R_k} A_{p,q} \gg \sum_{q \notin R_k} A_{p,q}, \quad \forall p \in R_k.
\end{equation}

\begin{figure*}[t]
  \centering
  \includegraphics[width=0.82\textwidth]{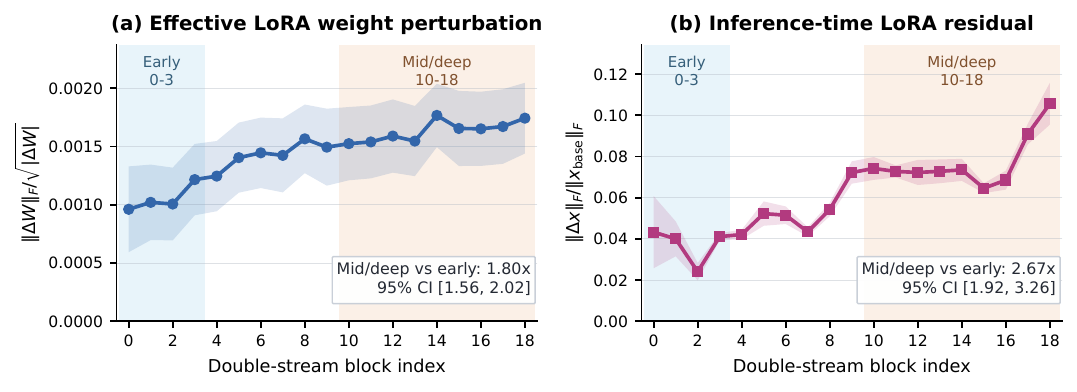}
  \captionsetup{width=0.95\linewidth}
  \caption{\rev{\textbf{LoRA perturbation magnitude across FLUX double-stream blocks.} Across all available LoRAs in the LoRAShop test set, early blocks carry substantially smaller LoRA perturbations than mid/deep blocks. Under the corresponding LoRA activation prompts, the inference-time LoRA residuals in mid/deep blocks are $2.67\times$ larger than in early blocks. Thus early global attention mainly mixes small LoRA residuals, while stronger LoRA effects emerge in more local semantic blocks.}}
  \label{fig:lora_perturbation_norms}
\end{figure*}

\rev{We further verify this mechanism by measuring LoRA perturbation magnitudes across FLUX double-stream blocks, as shown in Fig.~\ref{fig:lora_perturbation_norms}. The normalized effective LoRA weight perturbation is $1.80\times$ larger in mid/deep blocks than in early blocks, and the inference-time LoRA residual under the corresponding activation prompt is $2.67\times$ larger in mid/deep blocks. Therefore, although early blocks aggregate broader context, the LoRA-induced residuals being mixed there are empirically small; stronger LoRA effects emerge in mid-to-deep semantic blocks where attention locality is already much stronger. By applying spatial masks to the additive LoRA residual path, we effectively suppress direct cross-region LoRA interference while preserving the base model's global context propagation.}

\subsection{Exploring Intrinsic Segmentation Capabilities in Flow Models while Multi-Concept Generation}
\label{sec:seg}

\begin{figure*}[t]
  \centering
  \centering
  \includegraphics[width=1\textwidth]{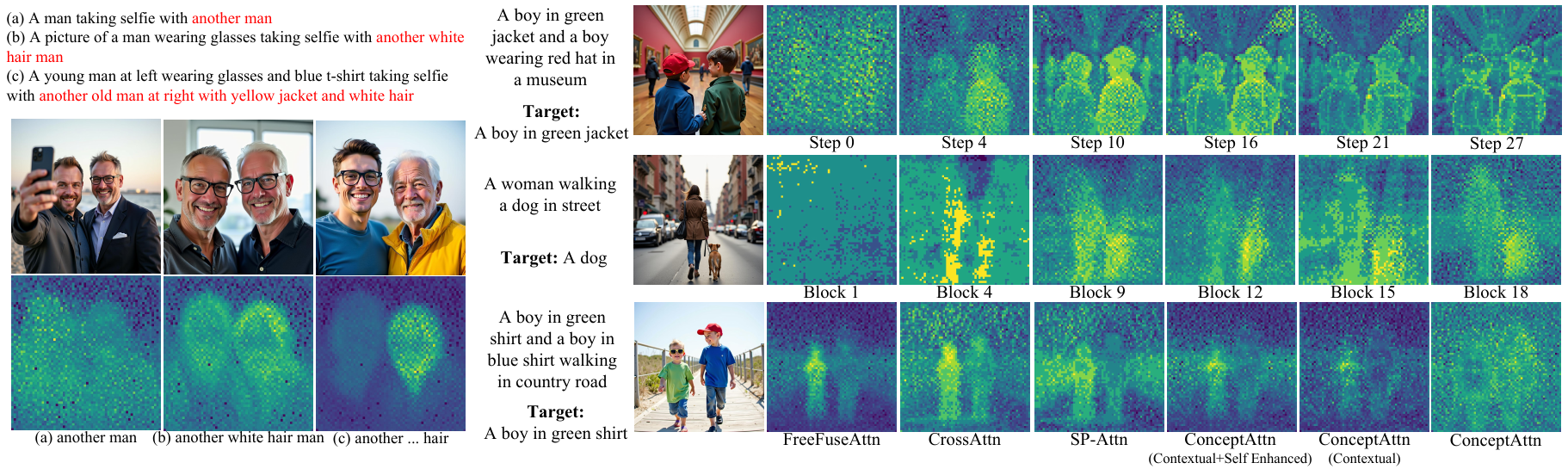}
  \captionsetup{width=1\linewidth}
  \caption{\textbf{Left}: High-quality descriptive prompts boost character distinguishability during generation and enhance similarity map quality. \textbf{Right}: Visualization of similarity maps across different dimensions. (a) Temporal Dynamics: Analyzing the cross attention heatmap in the denoising steps (top row) reveals that the alignment between text and image embeddings is most optimal during the early-to-mid stages. (b) Layer-wise Analysis: Within the Flux architecture (middle row), later blocks demonstrate better fusion of semantic and visual information. (c) Method Comparison: Compared to baseline methods (bottom row), our FreeFuse Attn exhibits the highest spatial discriminability, effectively disentangling symmetry concepts.}
  \label{fig:step_block_method_comp}
\end{figure*}


Prior literature~\citep{baranchuk2021label, tang2023daam, li2023openval, xu2023odiseopen, tian2024diffuseattend, helbling2025conceptattentiondiffusiontransformerslearn} has established that the latent representations of Diffusion and Flow Matching models exhibit superior semantic segmentation capabilities compared to discriminative baselines like CLIP or DINO. Whether achieved through post-hoc lightweight networks~\citep{xu2023odiseopen, li2023openval, baranchuk2021label} or by exploiting self-attention and cross-attention clustering~\citep{tang2023daam, helbling2025conceptattentiondiffusiontransformerslearn, tian2024diffuseattend}, these findings corroborate the robust intrinsic segmentation potential of generative models. Building upon this premise, we investigate how to efficiently localize multiple subjects within the Flux (representing Flow Matching DiT) for multi-subject generation. 

\begin{figure*}[t]
  \centering
  \includegraphics[width=0.82\textwidth]{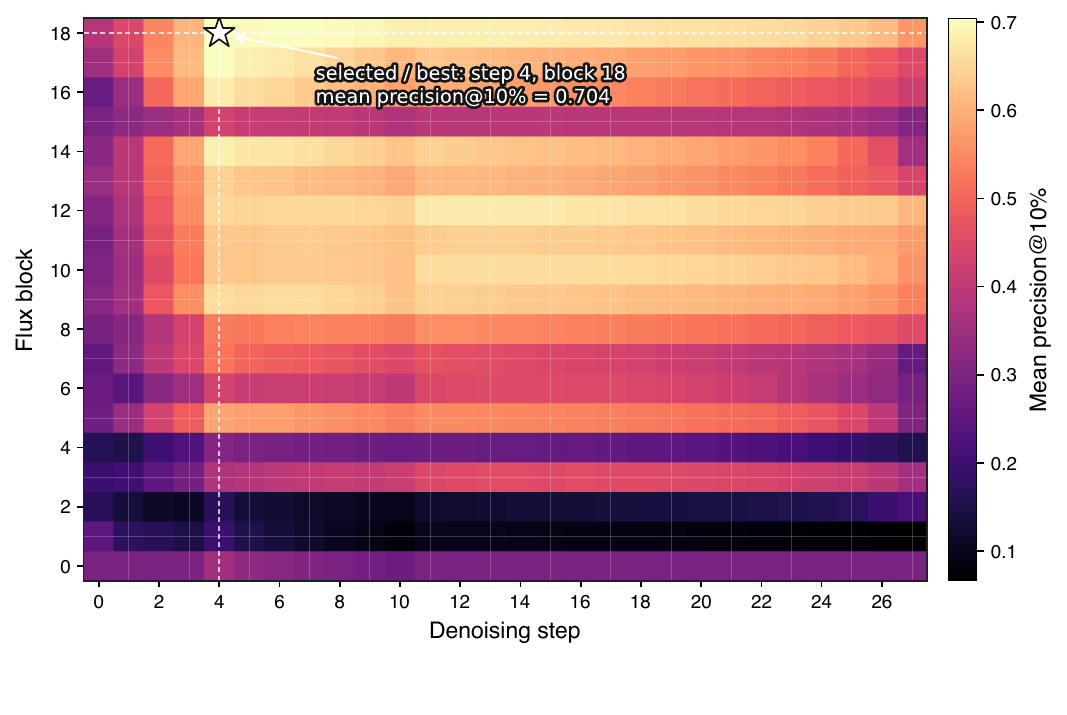}
  \captionsetup{width=0.95\linewidth}
  \caption{\rev{\textbf{Step/block sweep for FreeFuseAttn mask extraction.} We evaluate} \(\mathrm{Precision@10\%}\), \rev{defined as the percentage of the top 10\% activated tokens that fall inside the corresponding subject region, over 300 generated samples. The selected cell, block 18 at step index 4, achieves the best mean precision of 0.704, supporting our use of the last double-stream block at the 5th denoising step.}}
  \label{fig:step_block_sweep_heatmap}
\end{figure*}

We first determine the optimal temporal window and architectural layer for mask extraction. Through empirical analysis on Flux.dev, we identify the early-to-mid denoising steps as the optimal temporal window. As shown in right line 1 of Fig.~\ref{fig:step_block_method_comp}, we observe that at the initial timesteps, spatial structures remain nascent and indistinguishable from Gaussian noise; conversely, latents in later stages become dominated by high-frequency texture generation, showing a marked decoupling from semantic signals. This aligns with findings in~\citep{choi2022perception,tinaz2025emergence,qian2024boosting, lu2024coarse}, suggesting that the intermediate denoising phase represents the critical window where layout is established but not yet rigidified. \rev{We ultimately opted to extract the mask at the 5th step (step index 4 out of a total of 28 steps). We further evaluated semantic distinctiveness across different layers, pinpointing the output of the last Double Stream Block, \texttt{transformer\_blocks.18}, as the optimal locus for semantic-image alignment (see right line 2 of Fig.~\ref{fig:step_block_method_comp}). Fig.~\ref{fig:step_block_sweep_heatmap} further reports a quantitative step/block sweep under Precision@10\%, confirming that block 18 at step index 4 yields the strongest localization accuracy.} Comparative results for U-Net architectures (e.g., SDXL) are detailed in the Appendix.~\ref{sec:appendix_sdxl}. 

With the optimal spatiotemporal locus established, we scrutinized mainstream mask extraction paradigms, specifically comparison against the widely adopted Cross-Attention~\citep{meral2024clora,dalva2025lorashop,jiang2025mc} and the state-of-the-art ConceptAttn~\citep{helbling2025conceptattentiondiffusiontransformerslearn}. As depicted in the right line 3 of Fig.~\ref{fig:step_block_method_comp}, we identify a critical limitation in ConceptAttn: its independent encoding of concepts leads to semantic collapse when handling visually similar subjects, as the embeddings lack contextual discrimination. Although implementing a contextual variant (encoding all concepts jointly) partially alleviates this, our empirical analysis reveals that standard Cross-Attention still retains the highest raw semantic fidelity for subject alignment. However, Cross-Attention maps are spatially fragmented, suffering from the `hole phenomenon' identified in SPDiffusion~\citep{zhang2024spdiffusion}, where activations are sparse and non-target tokens exhibit significant noise. While SP-Attn~\citep{zhang2024spdiffusion} attempts to mitigate this, it fails to achieve sufficient spatial cohesiveness in multi-subject optimization. Formally, let $Q \in \mathbb{R}^{N \times d}$ denote the spatial query features and $K_c \in \mathbb{R}^{L_c \times d}$ denote the key features corresponding to the tokens of concept $c$ (where $L_c$ is the token count). For clarity, we present the formulation for a single attention head; in practice, the scores are averaged across all heads. Unlike standard cross-attention which normalizes over the textual dimension, we aim to obtain a spatial probability distribution for each concept token. We compute the spatial similarity map $\mathcal{A}_c \in \mathbb{R}^{N \times L_c}$ by applying Softmax along the spatial dimension $N$:
\begin{equation}
    \mathcal{A}_c = \text{Softmax}_{\text{spatial}}\left(\frac{Q K_c^\top}{\sqrt{d}}\right),
\end{equation}
where the entry $\mathcal{A}_c^{(p,l)}$ represents the contribution of the $p$-th image token to the $l$-th concept token, such that $\sum_{p=1}^N \mathcal{A}_c^{(p,l)} = 1$. To derive the aggregate activation map $S_c \in \mathbb{R}^N$ for the entire concept, we average the spatial responses across all tokens associated with concept $c$:
\begin{equation}
    S_c = \frac{1}{L_c} \sum_{l=1}^{L_c} \mathcal{A}_c[:, l].
\end{equation}

To identify distinct subject regions and suppress ambiguity, we compute the discriminative score $\hat{S}_c$. We enhance the signal by penalizing activations from competing concepts $j \neq c$:
\begin{equation}
    \hat{S}_c = M \cdot S_c - \sum_{j \neq c} S_j,
\end{equation}
where $M$ is the total number of concepts. We then identify the anchor set $\mathcal{P}_c = \text{TopK}(\hat{S}_c, k)$, pointing to the most representative spatial tokens. \rev{We use} \(k=\max(1,\lfloor0.1N\rfloor)\), \rev{i.e., the top 10\% spatial image tokens for each concept. For a} \(1024\times1024\) \rev{FLUX image,} \(N=(1024/16)^2=4096\), \rev{so} \(k=409\) \rev{anchors are selected.}

Finally, to generate a spatially cohesive mask $\mathcal{M}_c$ and mitigate the sparsity of raw cross-attention, we propagate semantic information from these anchors via latent similarity. \rev{Let} $Z\in\mathbb{R}^{N\times d}$ \rev{denote the image-token feature matrix from the selected FLUX block, and let} $Z_p \in \mathbb{R}^d$ \rev{denote the feature vector at anchor position} $p$:
\begin{equation}
    \mathcal{M}_c = \sigma \left( \frac{1}{\tau |\mathcal{P}_c|} \sum_{p \in \mathcal{P}_c} Z Z_p^\top \right).
\end{equation}
\rev{Here,} $ZZ_p^\top\in\mathbb{R}^{N}$ \rev{is a dense similarity map over image tokens,} $\sigma(\cdot)$ \rev{denotes a spatial normalization function (e.g, min-max scaling), and we set the temperature to} \(\tau=4000\). \rev{The resulting} \(\mathcal{M}_c\in\mathbb{R}^{N}\) \rev{is reshaped to the latent grid} \(H/16\times W/16\). This effectively reconstructs the dense subject shape from sparse anchor cues. We further conducted a systematic evaluation on 300 generated samples involving dual-subject interactions. We utilized the Segment Anything Model 3 (SAM3)~\citep{carion2025sam3} to generate ground-truth segmentation masks for each subject. We then computed the Precision@K metric, defined as the proportion of tokens falling within the ground-truth mask among the top $K\%$ of tokens with the highest similarity scores. As illustrated in Fig.~\ref{fig:locality_and_precision_curve} Right, FreeFuseAttn consistently outperforms baseline methods across all $K$ thresholds. Notably, while Cross-Attention suffers from low precision due to its sparse activation nature, and standard ConceptAttn collapses due to semantic ambiguity, our method achieves the highest alignment with the semantic ground truth. 

Additionally, empirical observations presented in Fig.~\ref{fig:step_block_method_comp} Left suggest that the efficacy of this reconstruction is positively correlated with the granularity of the textual descriptions. We find that supplementing subject prompts with distinctive attributes effectively mitigates the issue of map entanglement caused by semantic symmetry. This phenomenon aligns with intuition since richer semantic cues likely increase the orthogonality between concept embeddings in the latent space, thereby facilitating the identification of distinct anchors and yielding sharper similarity maps.

\subsection{Router Controlled and Bias-Guided Generation}

To preclude the mutual interference inherent in multi-LoRA activation, we perform an initial inference pass with all adapters deactivated. Leveraging the optimal spatiotemporal window and block identified in Sec.~\ref{sec:seg}, we extract similarity maps for each subject. Unlike~\citep{luo2024readoutguidancelearningcontrol, epstein2023diffusionselfguidancecontrollableimage, dalva2025lorashop} which relies on rigid thresholding for foreground separation, we explicitly utilize a background prompt to construct a competitive background similarity map, thereby significantly enhancing the precision of foreground-background delineation.

We subsequently refine these raw maps into robust binary masks using morphological opening and closing operations to ensure topological coherence. To resolve ambiguity in regions where subjects compete, we employ a contention resolution strategy incorporating territorial voting and gravity aggregation (refer to Appendix.~\ref{sec:appendix_postprocessing} for full formulation). 

With high-fidelity masks $\mathcal{M}$ established, we introduce the \textbf{Multi-LoRA Router Controlled and Bias Guided Generation} framework illustrated in Fig.~\ref{fig:method} stage 2. While prior approaches\citep{meral2024clora, sueyoshi2024predicated, rassin2023linguistic} attempt to mitigate the ``concept bleeding'' phenomenon via computationally expensive test-time optimization, our method directly leverages $\mathcal{M}$ to construct a spatial attention bias matrix. This matrix modulates the attention mechanism by encouraging image tokens to attend exclusively to their corresponding subject prompts while suppressing interactions with irrelevant subject descriptions. \rev{Concurrently, the Router enforces spatial locality by applying the adapter set $\mathcal{A}_s$ only to tokens inside the corresponding region $\mathcal{M}_s$. We define whole-image stylistic adjustment as the case where the user explicitly attaches a style LoRA to stylize the generated image rather than to modify a single localized identity or attribute, such as a Van Gogh painting style LoRA or a Miyazaki-style animation LoRA. Such a style adapter is handled by the same group-level rule: it is included in every foreground adapter set that should receive the style, i.e., $a_{\mathrm{style}}\in\mathcal{A}_s$ for all relevant $s$; when the background should share the same style, we assign the style adapter to the background or full-canvas region as well.}

\paragraph{\rev{Implementation scope.}}
\rev{For FLUX.1-dev, routing is applied to all LoRA-bearing transformer blocks: the 19 double-stream blocks \texttt{transformer\_blocks.0--18} and the 38 single-stream blocks \texttt{single\_transformer\_blocks.0--37}. In double-stream blocks, the Router gates image-token LoRA residuals in \texttt{to\_q}, \texttt{to\_k}, \texttt{to\_v}, \texttt{to\_out}, and the image feed-forward branch; subject text-token residuals are isolated in \texttt{add\_q\_proj}, \texttt{add\_k\_proj}, \texttt{add\_v\_proj}, \texttt{to\_add\_out}, and \texttt{ff\_context}. In single-stream blocks, routing is applied to the image-token LoRA residuals in \texttt{to\_q}, \texttt{to\_k}, \texttt{to\_v}, \texttt{proj\_mlp}, and \texttt{proj\_out}. During Phase 1, all adapters are disabled and masks are extracted from block 18 during the first $K=5$ of $T_{\mathrm{denoise}}=28$ steps. During Phase 2, we restart from the same initial latent and apply routed LoRA residuals at every denoising step. Importantly, routing gates only the additive LoRA residuals; the frozen base-model attention remains active, preserving global scene composition.}

\section{Experiments}

\paragraph{Experimental Setup.} To ensure a rigorous evaluation, we benchmark FreeFuse against five methods representing three distinct paradigms. (1) LoRA-based Fusion: We select LoRAShop~\citep{dalva2025lorashop} as the primary baseline. Crucially, we exclude older U-Net based methods to mitigate confounding variables arising from base model discrepancies, ensuring performance gains are attributable to our fusion algorithm rather than the Flux transformer's capabilities. (2) Modular Adapter Mechanisms: we focus on Multi-IP-Adapter and Multi-Redux as representative baselines. These exemplify the widely adopted "plug-and-play" paradigm, where lightweight modules inject visual conditions without altering base model weights. (3) Unified Native Models: We include OmniGen~\citep{xiao2025omnigen} and UMO~\citep{cheng2025umo}(UNO Based Version) to evaluate against emerging architectures designed with intrinsic multi-modal understanding. 

\subsection{Quantitative Results}

\begin{table*}[t]\small
\centering
\scriptsize{
\captionsetup{font=small}
\caption{{
Compared to existing methods, our approach demonstrates a significant advantage in Face Similarity and improvements in character and object consistency, while remaining on par with state-of-the-art models in instruction following and aesthetic evaluations.
}}
\label{tab:compare}
\resizebox{\linewidth}{!}{
\setlength\tabcolsep{2pt}
\renewcommand\arraystretch{1.1}
\begin{tabular}{r||cccc|cccc|cc|ccc|cc}
\hline
\hline
\rowcolor{gray!20}
 & 
\multicolumn{8}{c|}{\textbf{Reference Similarity Metrics}} & 
\multicolumn{2}{c|}{} &
\multicolumn{3}{c|}{} &
\multicolumn{2}{c}{}
\\
\cline{2-9}
\rowcolor{gray!20}
 & 
\multicolumn{4}{c|}{Character Similarity} & 
\multicolumn{4}{c|}{Object Similarity} & 
\multicolumn{2}{c|}{\multirow{-2}{*}{\textbf{Face Similarity Metrics}}} &
\multicolumn{3}{c|}{\multirow{-2}{*}{\shortstack{\textbf{Prompt Following and} \\ \textbf{Aesthetic Evaluation}}}} &
\multicolumn{2}{c}{\multirow{-2}{*}{\textbf{User Study}}}
\\
\rowcolor{gray!20}
\multirow{-3}{*}{Methods}  
& DINOv2$\uparrow$ & DINOv3$\uparrow$ & Dreamsim$\downarrow$ & Clip-I$\uparrow$
& DINOv2$\uparrow$ & DINOv3$\uparrow$ & Dreamsim$\downarrow$ & Clip-I$\uparrow$
& ArcFace$\uparrow$ & LVFace$\uparrow$
& Clip-T$\uparrow$ & HPSv2$\uparrow$ & HPSv3$\uparrow$
& Q1$\downarrow$ & Q2$\downarrow$
\\
\hline\hline
OmniGen 
& 0.4699 & \underline{0.5177} & 0.4233 & 0.6367
& 0.4662 & 0.5168 & 0.5048 & 0.6226
& 0.2990 & 0.1661
& 0.2088 & 0.2484 & 5.588
& 4.9 & 4.5
\\
\rowcolor{gray!10} UMO 
& 0.4378 & 0.4498 & 0.4710 & 0.5919
& \textbf{0.6535} & \textbf{0.6884} & \underline{0.3635} & \underline{0.7424}
& 0.3180 & 0.1661
& 0.2277 & 0.2629 & \textbf{8.756}
& 2.9 & 3.5
\\
Multi-Redux 
& 0.4433 & 0.4622 & 0.5363 & 0.5306
& 0.2456 & 0.2789 & 0.7255 & 0.3486
& 0.1475 & 0.0361
& 0.1225 & 0.2341 & 1.062
& 6.0 & 5.9
\\
\rowcolor{gray!10} Multi-IP-Adapter 
& \underline{0.4795} & 0.4918 & 0.4707 & 0.6028
& 0.4790 & 0.4970 & 0.5304 & 0.6424
& 0.1840 & 0.0710
& 0.2911 & 0.2831 & 7.197
& 3.9 & 3.4
\\
LoRAShop 
& 0.4597 & 0.4974 & \underline{0.4209} & \underline{0.6567}
& 0.6269 & 0.6394 & 0.4324 & 0.7129
& \underline{0.3886} & \underline{0.2350}
& \textbf{0.2980} & \textbf{0.2880} & \underline{8.494}
& \underline{2.2} & \underline{2.7}
\\
\hline
\rowcolor[HTML]{D7F6FF}
\textbf{Ours} 
& \textbf{0.4988} & \textbf{0.5235} & \textbf{0.3753} & \textbf{0.6764}
& \underline{0.6393} & \underline{0.6804} & \textbf{0.3516} & \textbf{0.7499}
& \textbf{0.4275} & \textbf{0.2534}
& \underline{0.2766} & \underline{0.2857} & 8.279
& \textbf{1.1} & \textbf{1.1}
\end{tabular}}}
\end{table*}

\begin{table*}[t]\small
\centering
\scriptsize{
\captionsetup{font=small}
\caption{Ablation study results indicating the effectiveness of each component.}
\label{tab:ablation}
\resizebox{\linewidth}{!}{
\setlength\tabcolsep{2pt}
\renewcommand\arraystretch{1.1}
\begin{tabular}{r||cccc|cccc|cc|ccc}
\hline
\hline
\rowcolor{gray!20}
 & 
\multicolumn{8}{c|}{\textbf{Reference Similarity Metrics}} & 
\multicolumn{2}{c|}{} &
\multicolumn{3}{c}{}
\\
\cline{2-9} 
\rowcolor{gray!20}
 & 
\multicolumn{4}{c|}{Character Similarity} & 
\multicolumn{4}{c|}{Object Similarity} & 
\multicolumn{2}{c|}{\multirow{-2}{*}{\textbf{Face Similarity Metrics}}} &
\multicolumn{3}{c}{\multirow{-2}{*}{\shortstack{\textbf{Prompt Following and} \\ \textbf{Aesthetic Evaluation}}}}
\\
\rowcolor{gray!20}
\multirow{-3}{*}{Methods}  
& DINOv2$\uparrow$ & DINOv3$\uparrow$ & Dreamsim$\downarrow$ & Clip-I$\uparrow$
& DINOv2$\uparrow$ & DINOv3$\uparrow$ & Dreamsim$\downarrow$ & Clip-I$\uparrow$
& ArcFace$\uparrow$ & LVFace$\uparrow$
& Clip-T$\uparrow$ & HPSv2$\uparrow$ & HPSv3$\uparrow$
\\
\hline\hline
Ours (Cross-Attn) 
& 0.4271 & 0.4378 & 0.4838 & 0.5919
& 0.6405 & 0.6832 & 0.3867 & 0.7378
& 0.3119 & 0.1975
& 0.2537 & 0.2762 & 6.704
\\
\rowcolor{gray!10} Ours (w/o Postprocessing) 
& \underline{0.4377} & \underline{0.4733} & 0.4303 & 0.6323
& 0.6055 & 0.6570 & 0.4052 & 0.7392
& 0.3259 & 0.1868
& \underline{0.2868} & 0.2846 & 7.727
\\
Ours (w/o Attn bias) 
& 0.4234 & 0.4691 & \underline{0.4186} & \underline{0.6448}
& \textbf{0.6589} & \underline{0.6939} & \underline{0.3522} & \textbf{0.7635}
& \underline{0.3491} & \underline{0.2027}
& \textbf{0.2882} & \textbf{0.2865} & \underline{7.971}
\\
\rowcolor{revblue} Ours (w/o Router; bias only)
& 0.3759 & 0.4138 & 0.5065 & 0.5855
& \underline{0.6495} & \textbf{0.7249} & 0.6546 & 0.7477
& 0.2669 & 0.1523
& 0.2678 & 0.2807 & 6.913
\\
\hline
\rowcolor[HTML]{D7F6FF}
\textbf{Ours(Full)} 
& \textbf{0.4988} & \textbf{0.5235} & \textbf{0.3753} & \textbf{0.6764}
& 0.6393 & 0.6804 & \textbf{0.3516} & \underline{0.7499}
& \textbf{0.4275} & \textbf{0.2534}
& 0.2766 & \underline{0.2857} & \textbf{8.279}
\\
\end{tabular}}}
\end{table*}

\begin{figure*}[t]
  \centering
  \centering
  \includegraphics[width=1\textwidth]{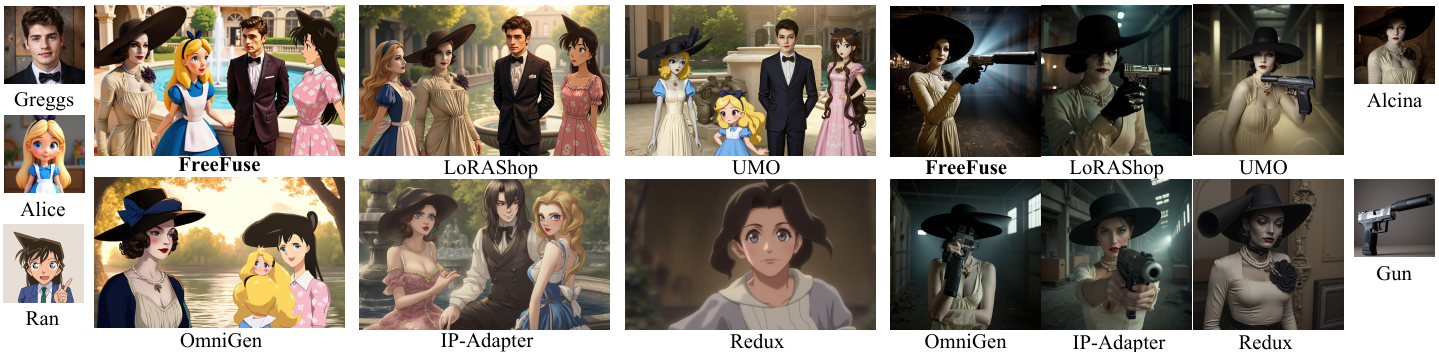}
  \captionsetup{width=1\linewidth}
  \caption{Qualitative Comparison. Prompt 1: \textit{<A>, <B>, <C>, and <D> are having a friendly conversation by the fountain pool}; Prompt 2: \textit{<A> is aiming a <Gun>.}}
  \label{fig:comp}
\end{figure*}

To rigorously evaluate the versatility of FreeFuse across diverse semantic domains, we curated a comprehensive multi-subject benchmark consisting of 15 distinct LoRAs. The subject set is strategically categorized to test robustness against domain shifts and varying spatial granularities: it includes 10 Character LoRAs spanning photorealistic humans, 2D anime figures, 3D avatars, and anthropomorphic entities, alongside 5 Object LoRAs ranging from large-scale vehicles to deformable apparel and intricate accessories. 

We push the evaluation complexity significantly beyond prior works\citep{kong2024omg,dalva2025lorashop,meral2024clora}, which typically limited assessment to dual-subject scenarios. Our protocol involves two distinct stress tests: (1) Multi-Character Interaction: We designed 20 distinct scenes with diverse backgrounds, each necessitating the simultaneous inference of 4 concurrent LoRAs, thereby imposing a high load on the fusion mechanism. (2) Character-Object Interaction: We synthesized composite prompts by conjoining character descriptions with specific object-interaction templates. In total, 470 images were generated for each method to ensure statistical significance. 

We assess performance using a multi-faceted metric suite that evaluates the generated results from three complementary perspectives: reference similarity between the generated images and the original character images, face similarity between each generated character and its corresponding ground-truth identity, and prompt following and aesthetic with respect to the given prompts. This allows us to comprehensively measure both character consistency and prompt-aligned generation quality. Specifically, we employ DINOv2~\citep{oquab2023dinov2} and DINOv3~\citep{simeoni2025dinov3} to measure coarse-grained visual fidelity, while CLIP-I~\citep{radford2021clip} and DreamSim~\citep{fu2023dreamsim} capture high-level semantic consistency. For stringent identity verification of human subjects, we incorporate ArcFace~\citep{deng2018arcface} and LVFace~\citep{you2025lvface}. Global instruction adherence is quantified via CLIP-T. For aesthetic quality, the evaluation is conducted using HPSv2~\citep{wu2023hpsv2} and HPSv3~\citep{ma2025hpsv3}. To validate perceptual alignment, we further conducted a user preference study (details in Appendix.~\ref{sec:user_study}).

As evidenced in Tab.~\ref{tab:compare}, FreeFuse establishes a new state-of-the-art in identity preservation. Our method demonstrates a commanding lead across both visual similarity metrics (DINOv2, DINOv3, Clip-I, Dreamsim) and fine-grained facial recognition benchmarks (ArcFace/LVFace). Furthermore, FreeFuse maintains superior object fidelity without compromising global image quality, achieving aesthetic scores (HPS) comparable to leading baselines while securing the top rank in human evaluations.

\begin{table}[t]
\centering
\small
\caption{\rev{Inference latency comparison (seconds) at $1024\times1024$ resolution on a single L20 GPU. FreeFuse adds a mask-extraction pass, but remains substantially faster than LoRAShop while avoiding the severe conflicts of naive multi-LoRA sampling.}}
\label{tab:latency_main}
\resizebox{\linewidth}{!}{%
\begin{tabular}{lccccc}
\toprule
\textbf{Method} & \textbf{2 LoRAs} & \textbf{3 LoRAs} & \textbf{4 LoRAs} & \textbf{5 LoRAs} & \textbf{6 LoRAs} \\
\midrule
\rowcolor{revblue}\textcolor{gray}{Naive Flux} & \textcolor{gray}{21.95} & \textcolor{gray}{24.60} & \textcolor{gray}{27.11} & \textcolor{gray}{29.64} & \textcolor{gray}{32.08} \\
\rowcolor{revblue}LoRAShop~\citep{dalva2025lorashop} & 64.78 & 84.36 & 103.60 & 122.96 & 142.27 \\
\rowcolor{revblue}\textbf{FreeFuse} & \textbf{36.08} & \textbf{42.96} & \textbf{50.29} & \textbf{58.63} & \textbf{67.96} \\
\midrule
\rowcolor{revblue}\textbf{Speedup vs. LoRAShop} & \textbf{1.80$\times$} & \textbf{1.96$\times$} & \textbf{2.06$\times$} & \textbf{2.10$\times$} & \textbf{2.09$\times$} \\
\bottomrule
\end{tabular}}
\end{table}

\subsection{Qualitative Results}

\begin{figure*}[t]
  \centering
  \centering
  \includegraphics[width=1\textwidth]{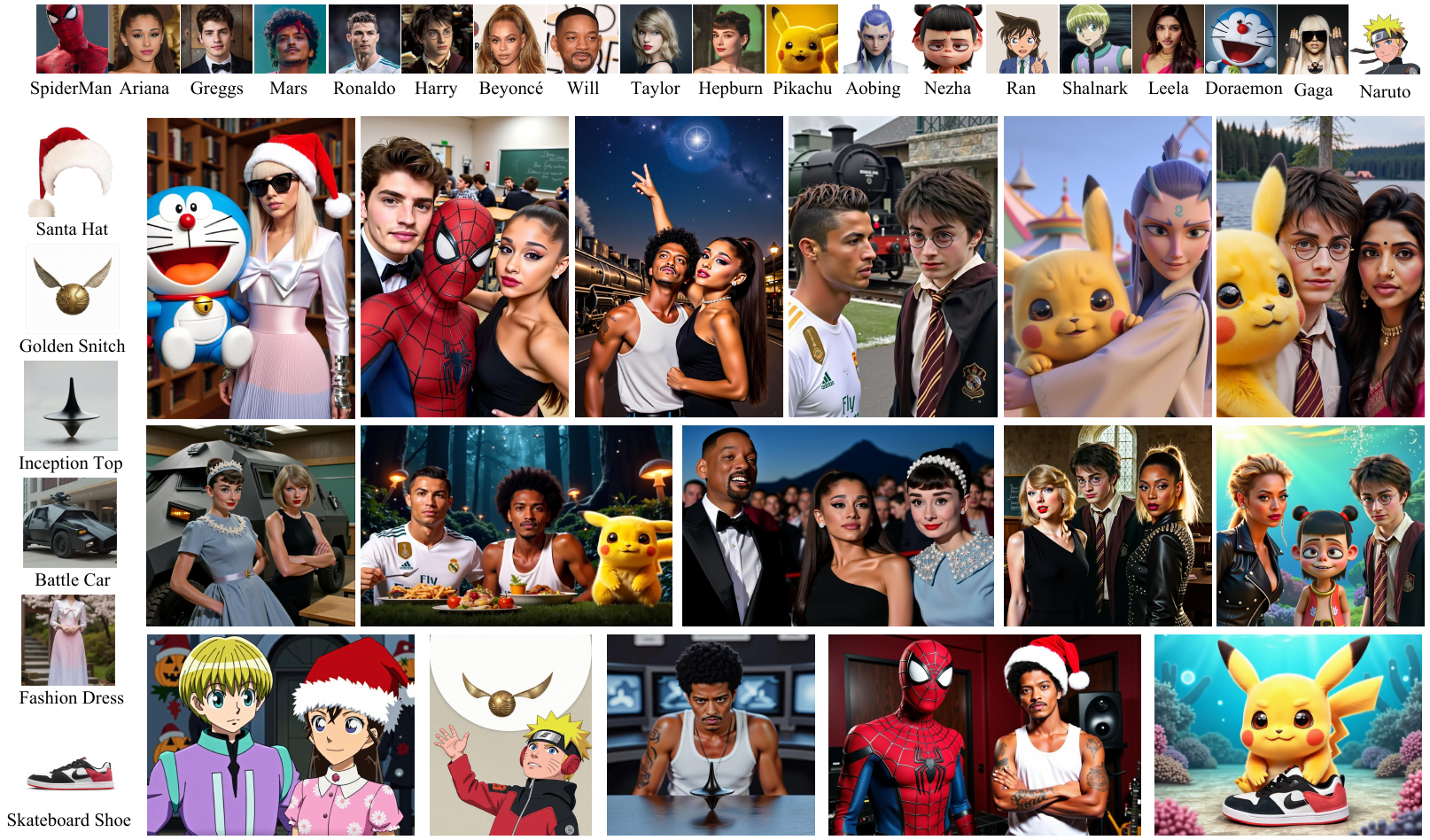}
  \captionsetup{width=1\linewidth}
  \caption{Additional visualization results. Our method achieves high-quality joint image generation across diverse categories, including real-world humans, anime/3D characters, non-human entities, garments, objects, and vehicles.}
  \label{fig:more_results}
\end{figure*}

Fig.~\ref{fig:comp} presents a visual comparison against baseline methods. Observe that competing approaches frequently suffer from severe feature leakage and identity blending, where attributes of one subject (e.g., skin tone, facial structure) erroneously bleed into another. In contrast, FreeFuse enforces semantic isolation, preserving the distinct identity of each subject even in spatially proximate interactions. To further validate versatility, Fig.~\ref{fig:more_results} displays a generation matrix involving 25 diverse characters and objects. Ranging from photorealistic humans to stylized anime figures and non-human entities, our method consistently synthesizes high-fidelity multi-subject interactions with coherent spatial layouts, demonstrating robust generalization across a broad semantic spectrum.

\subsection{Ablation Study}

To isolate the contribution of individual components, we conducted a systematic decomposition of the FreeFuse framework, with results summarized in Tab.~\ref{tab:ablation}. First, replacing our FreeFuseAttn with standard Cross-Attention yields a marked degradation in identity fidelity, confirming that raw attention maps are too sparse to support robust localization. Second, omitting the post-processing stage (morphological filtering and contention resolution) results in noisy masks; while general structure is maintained, the precision of boundary delineation suffers, leading to increased artifacts in complex interactions. \rev{Third, ablating the Bias-Guided Generation mechanism weakens semantic-spatial alignment and allows feature leakage to reappear. Finally, removing the Router while retaining only the attention bias produces a bias-only variant in which LoRA residuals are no longer spatially gated. Its substantial drop on character similarity and face identity metrics (e.g., ArcFace $0.2669$ vs. $0.4275$ for the full method) directly isolates the contribution of token-level LoRA routing.}

\topic{Discussion.}
A salient advantage of FreeFuse lies in its seamless modularity. By eschewing external segmentors and invasive weight updates, our framework preserves the original manifold of the base model. While primarily demonstrated on Flow Matching transformers, the core mechanism of FreeFuse is theoretically architecture-agnostic and generalizes to other backbones (e.g., SDXL), as detailed in Appendix.\ref{sec:appendix_sdxl}. This intrinsic design renders FreeFuse naturally compatible with downstream adapters: As illustrated in Fig.~\ref{fig:teaser}, it functions orthogonally to spatial guidance modules like ControlNet and reference encoders such as IP-Adapter or Redux. Consequently, FreeFuse not only resolves multi-subject conflicts but also unlocks fine-grained structural control and style transfer within complex compositional scenarios, expanding the applicability of pipelines.

\section{Limitations and Future Work}
\label{sec:limitation}

Our framework involves a two-stage pipeline, mask extraction followed by router-guided inference, which inevitably introduces computational overhead. However, \rev{as shown in Tab.~\ref{tab:latency_main} and detailed in Appendix~\ref{sec:latency_analysis}}, empirical benchmarks demonstrate that FreeFuse scales far more efficiently than existing training-free alternatives based on Flux (e.g., LoRAShop~\cite{dalva2025lorashop}). Future work will extend this mechanism to video generation, investigating how temporal attention dynamics can be leveraged to maintain multi-subject consistency across frames.

\section{Conclusion}
We present FreeFuse, a training-free framework for multi-subject LoRA fusion in text-to-image generation. FreeFuse mitigates feature conflicts by spatially routing each subject LoRA to its corresponding semantic region, without requiring LoRA retraining, auxiliary segmentation models, or user-specified masks. To obtain reliable routing regions, FreeFuseAttn leverages the intrinsic semantic alignment of diffusion transformers, combining cross-attention anchors with latent similarity propagation to produce dense and discriminative subject masks that alleviate sparse activations and hole artifacts. Based on these masks, adaptive token-level routing and bias-guided attention enforce local LoRA activation and suppress identity leakage during inference. Extensive experiments demonstrate that FreeFuse improves identity preservation, object consistency, and compositional fidelity while maintaining competitive prompt following and visual quality. 

\subsubsection*{Broader Impact Statement}
This research explores the technical possibilities of text-to-image compositional synthesis. We recognize the societal implications inherent in personalized generation technologies. The capability to integrate specific subject LoRAs carries the risk of misuse, including the creation of non-consensual content or misleading depictions (deepfakes). We strictly oppose the application of our method for generating NSFW content, harassment, or disinformation. We emphasize the importance of adhering to the safety guidelines and terms of use of the underlying foundational models. 


\subsubsection*{Acknowledgments}
This work was supported by National Natural Science Foundation of China (U23A20311, 62206245).

\bibliography{main}

@article{rassin2023linguistic,
  title={Linguistic binding in diffusion models: Enhancing attribute correspondence through attention map alignment},
  author={Rassin, Royi and Hirsch, Eran and Glickman, Daniel and Ravfogel, Shauli and Goldberg, Yoav and Chechik, Gal},
  journal={Advances in Neural Information Processing Systems},
  volume={36},
  pages={3536--3559},
  year={2023}
}

@inproceedings{sueyoshi2024predicated,
  title={Predicated diffusion: Predicate logic-based attention guidance for text-to-image diffusion models},
  author={Sueyoshi, Kota and Matsubara, Takashi},
  booktitle={Proceedings of the IEEE/CVF Conference on Computer Vision and Pattern Recognition},
  pages={8651--8660},
  year={2024}
}

@article{zhang2024spdiffusion,
  title={SPDiffusion: Semantic Protection Diffusion Models for Multi-concept Text-to-image Generation},
  author={Zhang, Yang and Zhang, Rui and Nie, Xuecheng and Li, Haochen and Chen, Jikun and Hao, Yifan and Zhang, Xin and Liu, Luoqi and Li, Ling},
  journal={arXiv preprint arXiv:2409.01327},
  year={2024}
}

@inproceedings{choi2022perception,
  title={Perception prioritized training of diffusion models},
  author={Choi, Jooyoung and Lee, Jungbeom and Shin, Chaehun and Kim, Sungwon and Kim, Hyunwoo and Yoon, Sungroh},
  booktitle={Proceedings of the IEEE/CVF conference on computer vision and pattern recognition},
  pages={11472--11481},
  year={2022}
}

@article{tinaz2025emergence,
  title={Emergence and Evolution of Interpretable Concepts in Diffusion Models},
  author={Tinaz, Berk and Fabian, Zalan and Soltanolkotabi, Mahdi},
  journal={arXiv preprint arXiv:2504.15473},
  year={2025}
}

@inproceedings{qian2024boosting,
  title={Boosting diffusion models with moving average sampling in frequency domain},
  author={Qian, Yurui and Cai, Qi and Pan, Yingwei and Li, Yehao and Yao, Ting and Sun, Qibin and Mei, Tao},
  booktitle={Proceedings of the IEEE/CVF conference on computer vision and pattern recognition},
  pages={8911--8920},
  year={2024}
}

@inproceedings{lu2024coarse,
  title={Coarse-to-fine latent diffusion for pose-guided person image synthesis},
  author={Lu, Yanzuo and Zhang, Manlin and Ma, Andy J and Xie, Xiaohua and Lai, Jianhuang},
  booktitle={Proceedings of the IEEE/CVF Conference on Computer Vision and Pattern Recognition},
  pages={6420--6429},
  year={2024}
}

@inproceedings{tang2023daam,
  title={What the daam: Interpreting stable diffusion using cross attention},
  author={Tang, Raphael and Liu, Linqing and Pandey, Akshat and Jiang, Zhiying and Yang, Gefei and Kumar, Karun and Stenetorp, Pontus and Lin, Jimmy and T{\"u}re, Ferhan},
  booktitle={Proceedings of the 61st Annual Meeting of the Association for Computational Linguistics (Volume 1: Long Papers)},
  pages={5644--5659},
  year={2023}
}

@inproceedings{xu2023odiseopen,
  title={Open-vocabulary panoptic segmentation with text-to-image diffusion models},
  author={Xu, Jiarui and Liu, Sifei and Vahdat, Arash and Byeon, Wonmin and Wang, Xiaolong and De Mello, Shalini},
  booktitle={Proceedings of the IEEE/CVF conference on computer vision and pattern recognition},
  pages={2955--2966},
  year={2023}
}

@article{baranchuk2021label,
  title={Label-efficient semantic segmentation with diffusion models},
  author={Baranchuk, Dmitry and Rubachev, Ivan and Voynov, Andrey and Khrulkov, Valentin and Babenko, Artem},
  journal={arXiv preprint arXiv:2112.03126},
  year={2021}
}

@inproceedings{tian2024diffuseattend,
  title={Diffuse attend and segment: Unsupervised zero-shot segmentation using stable diffusion},
  author={Tian, Junjiao and Aggarwal, Lavisha and Colaco, Andrea and Kira, Zsolt and Gonzalez-Franco, Mar},
  booktitle={Proceedings of the IEEE/CVF Conference on Computer Vision and Pattern Recognition},
  pages={3554--3563},
  year={2024}
}

@inproceedings{li2023openval,
  title={Open-vocabulary object segmentation with diffusion models},
  author={Li, Ziyi and Zhou, Qinye and Zhang, Xiaoyun and Zhang, Ya and Wang, Yanfeng and Xie, Weidi},
  booktitle={Proceedings of the IEEE/CVF International Conference on Computer Vision},
  pages={7667--7676},
  year={2023}
}

@misc{dalva2025lorashop,
  title={LoRAShop: Training-Free Multi-Concept Image Generation and Editing with Rectified Flow Transformers}, 
  author={Yusuf Dalva and Hidir Yesiltepe and Pinar Yanardag},
  year={2025},
  eprint={2505.23758},
  archivePrefix={arXiv},
  primaryClass={cs.CV},
  url={https://arxiv.org/abs/2505.23758}, 
}

@inproceedings{simsar2025loraclr,
  title={LoRACLR: Contrastive Adaptation for Customization of Diffusion Models},
  author={Simsar, Enis and Hofmann, Thomas and Tombari, Federico and Yanardag, Pinar},
  booktitle={Proceedings of the Computer Vision and Pattern Recognition Conference},
  pages={13189--13198},
  year={2025}
}

@inproceedings{jiang2025mc,
  title={MC\^{} 2: Multi-concept Guidance for Customized Multi-concept Generation},
  author={Jiang, Jiaxiu and Zhang, Yabo and Feng, Kailai and Wu, Xiaohe and Li, Wenbo and Pei, Renjing and Li, Fan and Zuo, Wangmeng},
  booktitle={Proceedings of the Computer Vision and Pattern Recognition Conference},
  pages={2802--2812},
  year={2025}
}

@article{woo2025flipconcept,
  title={Flipconcept: Tuning-free multi-concept personalization for text-to-image generation},
  author={Woo, Young Beom and Kim, Sun Eung and Lee, Seong-Whan},
  journal={arXiv preprint arXiv:2502.15203},
  year={2025}
}

@article{cheng2025umo,
  title={UMO: Scaling Multi-Identity Consistency for Image Customization via Matching Reward},
  author={Cheng, Yufeng and Wu, Wenxu and Wu, Shaojin and Huang, Mengqi and Ding, Fei and He, Qian},
  journal={arXiv preprint arXiv:2509.06818},
  year={2025}
}

@article{wu2025uno,
  title={Less-to-more generalization: Unlocking more controllability by in-context generation},
  author={Wu, Shaojin and Huang, Mengqi and Wu, Wenxu and Cheng, Yufeng and Ding, Fei and He, Qian},
  journal={arXiv preprint arXiv:2504.02160},
  year={2025}
}

@inproceedings{xiao2025omnigen,
  title={Omnigen: Unified image generation},
  author={Xiao, Shitao and Wang, Yueze and Zhou, Junjie and Yuan, Huaying and Xing, Xingrun and Yan, Ruiran and Li, Chaofan and Wang, Shuting and Huang, Tiejun and Liu, Zheng},
  booktitle={Proceedings of the Computer Vision and Pattern Recognition Conference},
  pages={13294--13304},
  year={2025}
}

@article{chen2025xverse,
  title={XVerse: Consistent Multi-Subject Control of Identity and Semantic Attributes via DiT Modulation},
  author={Chen, Bowen and Zhao, Mengyi and Sun, Haomiao and Chen, Li and Wang, Xu and Du, Kang and Wu, Xinglong},
  journal={arXiv preprint arXiv:2506.21416},
  year={2025}
}

@article{garibi2025tokenverse,
  title={Tokenverse: Versatile multi-concept personalization in token modulation space},
  author={Garibi, Daniel and Yadin, Shahar and Paiss, Roni and Tov, Omer and Zada, Shiran and Ephrat, Ariel and Michaeli, Tomer and Mosseri, Inbar and Dekel, Tali},
  journal={ACM Transactions On Graphics (TOG)},
  volume={44},
  number={4},
  pages={1--11},
  year={2025},
  publisher={ACM New York, NY, USA}
}

@misc{raghu2022visiontransformerslikeconvolutional,
      title={Do Vision Transformers See Like Convolutional Neural Networks?}, 
      author={Maithra Raghu and Thomas Unterthiner and Simon Kornblith and Chiyuan Zhang and Alexey Dosovitskiy},
      year={2022},
      eprint={2108.08810},
      archivePrefix={arXiv},
      primaryClass={cs.CV},
      url={https://arxiv.org/abs/2108.08810}, 
}

@misc{caron2021emergingpropertiesselfsupervisedvision,
      title={Emerging Properties in Self-Supervised Vision Transformers}, 
      author={Mathilde Caron and Hugo Touvron and Ishan Misra and Hervé Jégou and Julien Mairal and Piotr Bojanowski and Armand Joulin},
      year={2021},
      eprint={2104.14294},
      archivePrefix={arXiv},
      primaryClass={cs.CV},
      url={https://arxiv.org/abs/2104.14294}, 
}

@misc{helbling2025conceptattentiondiffusiontransformerslearn,
      title={ConceptAttention: Diffusion Transformers Learn Highly Interpretable Features}, 
      author={Alec Helbling and Tuna Han Salih Meral and Ben Hoover and Pinar Yanardag and Duen Horng Chau},
      year={2025},
      eprint={2502.04320},
      archivePrefix={arXiv},
      primaryClass={cs.CV},
      url={https://arxiv.org/abs/2502.04320}, 
}

@misc{shi2025dreamrelationbridgingcustomizationrelation,
      title={DreamRelation: Bridging Customization and Relation Generation}, 
      author={Qingyu Shi and Lu Qi and Jianzong Wu and Jinbin Bai and Jingbo Wang and Yunhai Tong and Xiangtai Li},
      year={2025},
      eprint={2410.23280},
      archivePrefix={arXiv},
      primaryClass={cs.CV},
      url={https://arxiv.org/abs/2410.23280}, 
}

@misc{luo2024readoutguidancelearningcontrol,
      title={Readout Guidance: Learning Control from Diffusion Features}, 
      author={Grace Luo and Trevor Darrell and Oliver Wang and Dan B Goldman and Aleksander Holynski},
      year={2024},
      eprint={2312.02150},
      archivePrefix={arXiv},
      primaryClass={cs.CV},
      url={https://arxiv.org/abs/2312.02150}, 
}

@misc{epstein2023diffusionselfguidancecontrollableimage,
      title={Diffusion Self-Guidance for Controllable Image Generation}, 
      author={Dave Epstein and Allan Jabri and Ben Poole and Alexei A. Efros and Aleksander Holynski},
      year={2023},
      eprint={2306.00986},
      archivePrefix={arXiv},
      primaryClass={cs.CV},
      url={https://arxiv.org/abs/2306.00986}, 
}

@misc{po2024orthogonaladaptationmodularcustomization,
      title={Orthogonal Adaptation for Modular Customization of Diffusion Models}, 
      author={Ryan Po and Guandao Yang and Kfir Aberman and Gordon Wetzstein},
      year={2024},
      eprint={2312.02432},
      archivePrefix={arXiv},
      primaryClass={cs.CV},
      url={https://arxiv.org/abs/2312.02432}, 
}

@article{ma2025hpsv3,
  title={Hpsv3: Towards wide-spectrum human preference score},
  author={Ma, Yuhang and Shui, Yunhao and Wu, Xiaoshi and Sun, Keqiang and Li, Hongsheng},
  journal={arXiv preprint arXiv:2508.03789},
  year={2025}
}

@article{wu2023hpsv2,
  title={Human preference score v2: A solid benchmark for evaluating human preferences of text-to-image synthesis},
  author={Wu, Xiaoshi and Hao, Yiming and Sun, Keqiang and Chen, Yixiong and Zhu, Feng and Zhao, Rui and Li, Hongsheng},
  journal={arXiv preprint arXiv:2306.09341},
  year={2023}
}

@article{carion2025sam3,
  title={Sam 3: Segment anything with concepts},
  author={Carion, Nicolas and Gustafson, Laura and Hu, Yuan-Ting and Debnath, Shoubhik and Hu, Ronghang and Suris, Didac and Ryali, Chaitanya and Alwala, Kalyan Vasudev and Khedr, Haitham and Huang, Andrew and others},
  journal={arXiv preprint arXiv:2511.16719},
  year={2025}
}

@article{fu2023dreamsim,
  title={Dreamsim: Learning new dimensions of human visual similarity using synthetic data},
  author={Fu, Stephanie and Tamir, Netanel and Sundaram, Shobhita and Chai, Lucy and Zhang, Richard and Dekel, Tali and Isola, Phillip},
  journal={arXiv preprint arXiv:2306.09344},
  year={2023}
}

@article{simeoni2025dinov3,
  title={Dinov3},
  author={Sim{\'e}oni, Oriane and Vo, Huy V and Seitzer, Maximilian and Baldassarre, Federico and Oquab, Maxime and Jose, Cijo and Khalidov, Vasil and Szafraniec, Marc and Yi, Seungeun and Ramamonjisoa, Micha{\"e}l and others},
  journal={arXiv preprint arXiv:2508.10104},
  year={2025}
}

@article{oquab2023dinov2,
  title={Dinov2: Learning robust visual features without supervision},
  author={Oquab, Maxime and Darcet, Timoth{\'e}e and Moutakanni, Th{\'e}o and Vo, Huy and Szafraniec, Marc and Khalidov, Vasil and Fernandez, Pierre and Haziza, Daniel and Massa, Francisco and El-Nouby, Alaaeldin and others},
  journal={arXiv preprint arXiv:2304.07193},
  year={2023}
}

@inproceedings{you2025lvface,
  title={{LVFace}: Progressive Cluster Optimization for Large Vision Models in Face Recognition},
  author={You, Jinghan and Li, Shanglin and Sun, Yuanrui and Wei, Jiangchuan and Guo, Mingyu and Feng, Chao and Ran, Jiao},
  booktitle={ICCV},
  year={2025}
}

@inproceedings{deng2018arcface,
title={ArcFace: Additive Angular Margin Loss for Deep Face Recognition},
author={Deng, Jiankang and Guo, Jia and Niannan, Xue and Zafeiriou, Stefanos},
booktitle={CVPR},
year={2019}
}

@article{meral2024clora,
  title={Clora: A contrastive approach to compose multiple lora models},
  author={Meral, Tuna Han Salih and Simsar, Enis and Tombari, Federico and Yanardag, Pinar},
  journal={arXiv preprint arXiv:2403.19776},
  year={2024}
}

@inproceedings{kwon2024concept,
  title={Concept weaver: Enabling multi-concept fusion in text-to-image models},
  author={Kwon, Gihyun and Jenni, Simon and Li, Dingzeyu and Lee, Joon-Young and Ye, Jong Chul and Heilbron, Fabian Caba},
  booktitle={Proceedings of the IEEE/CVF Conference on Computer Vision and Pattern Recognition},
  pages={8880--8889},
  year={2024}
}

@article{zhong2024multi,
  title={Multi-lora composition for image generation},
  author={Zhong, Ming and Shen, Yelong and Wang, Shuohang and Lu, Yadong and Jiao, Yizhu and Ouyang, Siru and Yu, Donghan and Han, Jiawei and Chen, Weizhu},
  journal={arXiv preprint arXiv:2402.16843},
  year={2024}
}

@inproceedings{kong2024omg,
  title={Omg: Occlusion-friendly personalized multi-concept generation in diffusion models},
  author={Kong, Zhe and Zhang, Yong and Yang, Tianyu and Wang, Tao and Zhang, Kaihao and Wu, Bizhu and Chen, Guanying and Liu, Wei and Luo, Wenhan},
  booktitle={European Conference on Computer Vision},
  pages={253--270},
  year={2024},
  organization={Springer}
}

@article{ouyang2025k,
  title={K-lora: Unlocking training-free fusion of any subject and style loras},
  author={Ouyang, Ziheng and Li, Zhen and Hou, Qibin},
  journal={arXiv preprint arXiv:2502.18461},
  year={2025}
}

@inproceedings{shah2024ziplora,
  title={Ziplora: Any subject in any style by effectively merging loras},
  author={Shah, Viraj and Ruiz, Nataniel and Cole, Forrester and Lu, Erika and Lazebnik, Svetlana and Li, Yuanzhen and Jampani, Varun},
  booktitle={European Conference on Computer Vision},
  pages={422--438},
  year={2024},
  organization={Springer}
}

@article{wu2025sd,
  title={Sd-lora: Scalable decoupled low-rank adaptation for class incremental learning},
  author={Wu, Yichen and Piao, Hongming and Huang, Long-Kai and Wang, Renzhen and Li, Wanhua and Pfister, Hanspeter and Meng, Deyu and Ma, Kede and Wei, Ying},
  journal={arXiv preprint arXiv:2501.13198},
  year={2025}
}

@article{dettmers2023qlora,
  title={Qlora: Efficient finetuning of quantized llms},
  author={Dettmers, Tim and Pagnoni, Artidoro and Holtzman, Ari and Zettlemoyer, Luke},
  journal={Advances in neural information processing systems},
  volume={36},
  pages={10088--10115},
  year={2023}
}

@article{gu2023mix,
  title={Mix-of-show: Decentralized low-rank adaptation for multi-concept customization of diffusion models},
  author={Gu, Yuchao and Wang, Xintao and Wu, Jay Zhangjie and Shi, Yujun and Chen, Yunpeng and Fan, Zihan and Xiao, Wuyou and Zhao, Rui and Chang, Shuning and Wu, Weijia and others},
  journal={Advances in Neural Information Processing Systems},
  volume={36},
  pages={15890--15902},
  year={2023}
}

@inproceedings{radford2021clip,
  title={Learning transferable visual models from natural language supervision},
  author={Radford, Alec and Kim, Jong Wook and Hallacy, Chris and Ramesh, Aditya and Goh, Gabriel and Agarwal, Sandhini and Sastry, Girish and Askell, Amanda and Mishkin, Pamela and Clark, Jack and others},
  booktitle={International conference on machine learning},
  pages={8748--8763},
  year={2021},
  organization={PmLR}
}

@inproceedings{yeh2023navigating,
  title={Navigating text-to-image customization: From lycoris fine-tuning to model evaluation},
  author={Yeh, Shih-Ying and Hsieh, Yu-Guan and Gao, Zhidong and Yang, Bernard BW and Oh, Giyeong and Gong, Yanmin},
  booktitle={The Twelfth International Conference on Learning Representations},
  year={2023}
}

@inproceedings{ruiz2023dreambooth,
  title={Dreambooth: Fine tuning text-to-image diffusion models for subject-driven generation},
  author={Ruiz, Nataniel and Li, Yuanzhen and Jampani, Varun and Pritch, Yael and Rubinstein, Michael and Aberman, Kfir},
  booktitle={Proceedings of the IEEE/CVF conference on computer vision and pattern recognition},
  pages={22500--22510},
  year={2023}
}

@article{wang2024instantid,
  title={Instantid: Zero-shot identity-preserving generation in seconds},
  author={Wang, Qixun and Bai, Xu and Wang, Haofan and Qin, Zekui and Chen, Anthony and Li, Huaxia and Tang, Xu and Hu, Yao},
  journal={arXiv preprint arXiv:2401.07519},
  year={2024}
}

@article{ye2023ip,
  title={Ip-adapter: Text compatible image prompt adapter for text-to-image diffusion models},
  author={Ye, Hu and Zhang, Jun and Liu, Sibo and Han, Xiao and Yang, Wei},
  journal={arXiv preprint arXiv:2308.06721},
  year={2023}
}

@article{gal2022image,
  title={An image is worth one word: Personalizing text-to-image generation using textual inversion},
  author={Gal, Rinon and Alaluf, Yuval and Atzmon, Yuval and Patashnik, Or and Bermano, Amit H and Chechik, Gal and Cohen-Or, Daniel},
  journal={arXiv preprint arXiv:2208.01618},
  year={2022}
}

@article{wu2025qwen,
  title={Qwen-image technical report},
  author={Wu, Chenfei and Li, Jiahao and Zhou, Jingren and Lin, Junyang and Gao, Kaiyuan and Yan, Kun and Yin, Sheng-ming and Bai, Shuai and Xu, Xiao and Chen, Yilei and others},
  journal={arXiv preprint arXiv:2508.02324},
  year={2025}
}

@inproceedings{esser2024scaling,
  title={Scaling rectified flow transformers for high-resolution image synthesis},
  author={Esser, Patrick and Kulal, Sumith and Blattmann, Andreas and Entezari, Rahim and M{\"u}ller, Jonas and Saini, Harry and Levi, Yam and Lorenz, Dominik and Sauer, Axel and Boesel, Frederic and others},
  booktitle={Forty-first international conference on machine learning},
  year={2024}
}

@article{podell2023sdxl,
  title={Sdxl: Improving latent diffusion models for high-resolution image synthesis},
  author={Podell, Dustin and English, Zion and Lacey, Kyle and Blattmann, Andreas and Dockhorn, Tim and M{\"u}ller, Jonas and Penna, Joe and Rombach, Robin},
  journal={arXiv preprint arXiv:2307.01952},
  year={2023}
}

@inproceedings{peebles2023scalable,
  title={Scalable diffusion models with transformers},
  author={Peebles, William and Xie, Saining},
  booktitle={Proceedings of the IEEE/CVF international conference on computer vision},
  pages={4195--4205},
  year={2023}
}

@inproceedings{rombach2022high,
  title={High-resolution image synthesis with latent diffusion models},
  author={Rombach, Robin and Blattmann, Andreas and Lorenz, Dominik and Esser, Patrick and Ommer, Bj{\"o}rn},
  booktitle={Proceedings of the IEEE/CVF conference on computer vision and pattern recognition},
  pages={10684--10695},
  year={2022}
}

@article{song2020denoising,
  title={Denoising diffusion implicit models},
  author={Song, Jiaming and Meng, Chenlin and Ermon, Stefano},
  journal={arXiv preprint arXiv:2010.02502},
  year={2020}
}

@article{ho2020denoising,
  title={Denoising diffusion probabilistic models},
  author={Ho, Jonathan and Jain, Ajay and Abbeel, Pieter},
  journal={Advances in neural information processing systems},
  volume={33},
  pages={6840--6851},
  year={2020}
}

@inproceedings{ronneberger2015u,
  title={U-net: Convolutional networks for biomedical image segmentation},
  author={Ronneberger, Olaf and Fischer, Philipp and Brox, Thomas},
  booktitle={International Conference on Medical image computing and computer-assisted intervention},
  pages={234--241},
  year={2015},
  organization={Springer}
}

@inproceedings{karras2019style,
  title={A style-based generator architecture for generative adversarial networks},
  author={Karras, Tero and Laine, Samuli and Aila, Timo},
  booktitle={Proceedings of the IEEE/CVF conference on computer vision and pattern recognition},
  pages={4401--4410},
  year={2019}
}

@inproceedings{karras2020analyzing,
  title={Analyzing and improving the image quality of stylegan},
  author={Karras, Tero and Laine, Samuli and Aittala, Miika and Hellsten, Janne and Lehtinen, Jaakko and Aila, Timo},
  booktitle={Proceedings of the IEEE/CVF conference on computer vision and pattern recognition},
  pages={8110--8119},
  year={2020}
}

@misc{flux2024,
    author={Black Forest Labs},
    title={FLUX},
    year={2024},
    howpublished={\url{https://github.com/black-forest-labs/flux}},
}

@inproceedings{arjovsky2017wasserstein,
  title={Wasserstein generative adversarial networks},
  author={Arjovsky, Martin and Chintala, Soumith and Bottou, L{\'e}on},
  booktitle={International conference on machine learning},
  pages={214--223},
  year={2017},
  organization={PMLR}
}

@article{goodfellow2014generative,
  title={Generative adversarial nets},
  author={Goodfellow, Ian J and Pouget-Abadie, Jean and Mirza, Mehdi and Xu, Bing and Warde-Farley, David and Ozair, Sherjil and Courville, Aaron and Bengio, Yoshua},
  journal={Advances in neural information processing systems},
  volume={27},
  year={2014}
}

@misc{labs2025flux1kontextflowmatching,
      title={FLUX.1 Kontext: Flow Matching for In-Context Image Generation and Editing in Latent Space},
      author={Black Forest Labs and Stephen Batifol and Andreas Blattmann and Frederic Boesel and Saksham Consul and Cyril Diagne and Tim Dockhorn and Jack English and Zion English and Patrick Esser and Sumith Kulal and Kyle Lacey and Yam Levi and Cheng Li and Dominik Lorenz and Jonas Müller and Dustin Podell and Robin Rombach and Harry Saini and Axel Sauer and Luke Smith},
      year={2025},
      eprint={2506.15742},
      archivePrefix={arXiv},
      primaryClass={cs.GR},
      url={https://arxiv.org/abs/2506.15742},
}

@article{cai2025hidream,
  title={HiDream-I1: A High-Efficient Image Generative Foundation Model with Sparse Diffusion Transformer},
  author={Cai, Qi and Chen, Jingwen and Chen, Yang and Li, Yehao and Long, Fuchen and Pan, Yingwei and Qiu, Zhaofan and Zhang, Yiheng and Gao, Fengbin and Xu, Peihan and others},
  journal={arXiv preprint arXiv:2505.22705},
  year={2025}
}

@article{hu2022lora,
  title={Lora: Low-rank adaptation of large language models.},
  author={Hu, Edward J and Shen, Yelong and Wallis, Phillip and Allen-Zhu, Zeyuan and Li, Yuanzhi and Wang, Shean and Wang, Lu and Chen, Weizhu and others},
  journal={ICLR},
  volume={1},
  number={2},
  pages={3},
  year={2022}
}
\bibliographystyle{tmlr}

\appendix
\section{Generalizability to U-Net Architectures (SDXL)}
\label{sec:appendix_sdxl}

While our primary investigation focuses on the Flow Matching architecture (specifically FLUX.1-dev~\cite{flux2024}) due to its superior prompt adherence and image quality, the core mechanism of FreeFuse, specifically the \textit{FreeFuseAttn} for mask extraction, the \textit{Token-Level Routing} for inference control and the \textit{Bias-Guided Generation}, is theoretically architecture-agnostic. In this section, we validate the extensibility of our framework by applying it to SDXL~\cite{podell2023sdxl}, a representative U-Net-based diffusion model.

\begin{figure*}[h]
  \centering
  \centering
  \includegraphics[width=1\textwidth]{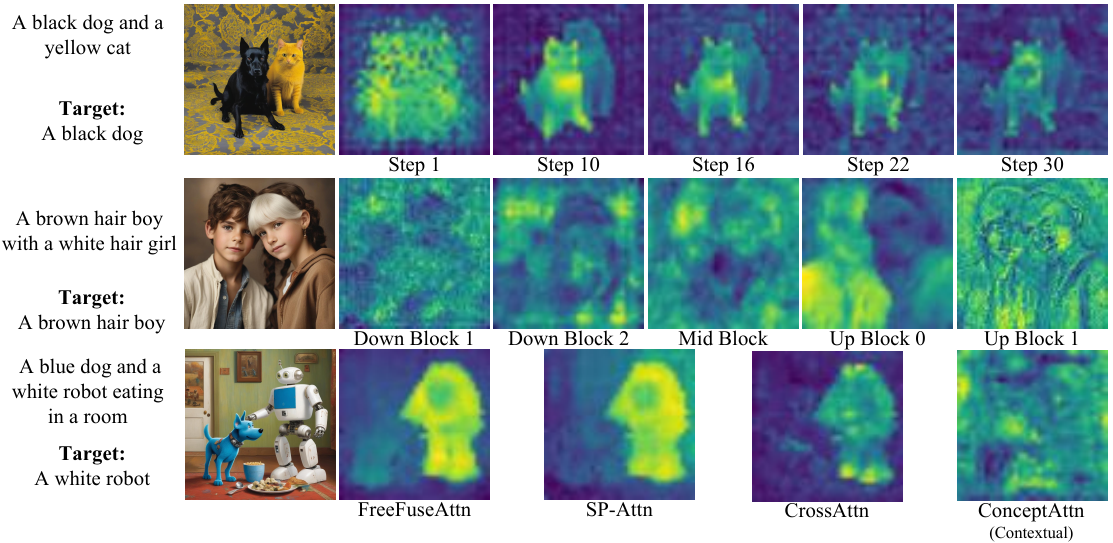}
  \captionsetup{width=1\linewidth}
  \caption{Visualization of similarity maps across different dimensions. (a) Temporal Dynamics: Analyzing the cross attention heatmap in the denoising steps (top row) reveals that the alignment between text and image embeddings is most optimal during the early-to-mid stages. (b) Layer-wise Analysis: Within the SDXL architecture (middle row), up block 0 demonstrate better fusion of semantic and visual information. (c) Method Comparison: Compared to baseline methods (bottom row), our FreeFuse Attn exhibits the highest spatial discriminability, effectively disentangling symmetry concepts.}
  \label{fig:sdxl_choices}
\end{figure*}

\subsection{Implementation Details on SDXL}
Unlike the unified DiT blocks in Flux, SDXL employs a separation of cross-attention and self-attention within the U-Net backbone. To adapt FreeFuse to this architecture, we implemented the following adjustments:

\begin{itemize}
    \item \textbf{Layer Selection:} We extract attention maps from the mid-level cross-attention layers of the U-Net decoder (specifically, \texttt{up\_blocks.0.attentions.0.transformer\_blocks.3.attn2}). Empirical analysis suggests these layers contain the most semantic-rich spatial layout information necessary for mask generation.
    \item \textbf{Routing Application:} The token-level routing is applied to the corresponding self-attention layers, enforcing the exclusivity constraint defined in the main paper. \rev{This gates direct LoRA residual injection across conflicting subject regions while preserving the base model's pretrained attention pathway.}
    \item \textbf{Bias-Guided Generation:} We extend the attention bias mechanism to the U-Net's cross-attention layers. By injecting a mask-derived bias matrix into the cross-attention scores, we actively suppress the attention weights of irrelevant subject prompts outside their designated regions (as defined by the masks extracted in the first stage), thereby minimizing semantic leakage.
\end{itemize}

\begin{figure}[t]
  \centering
  \centering
  \includegraphics[width=0.65\textwidth]{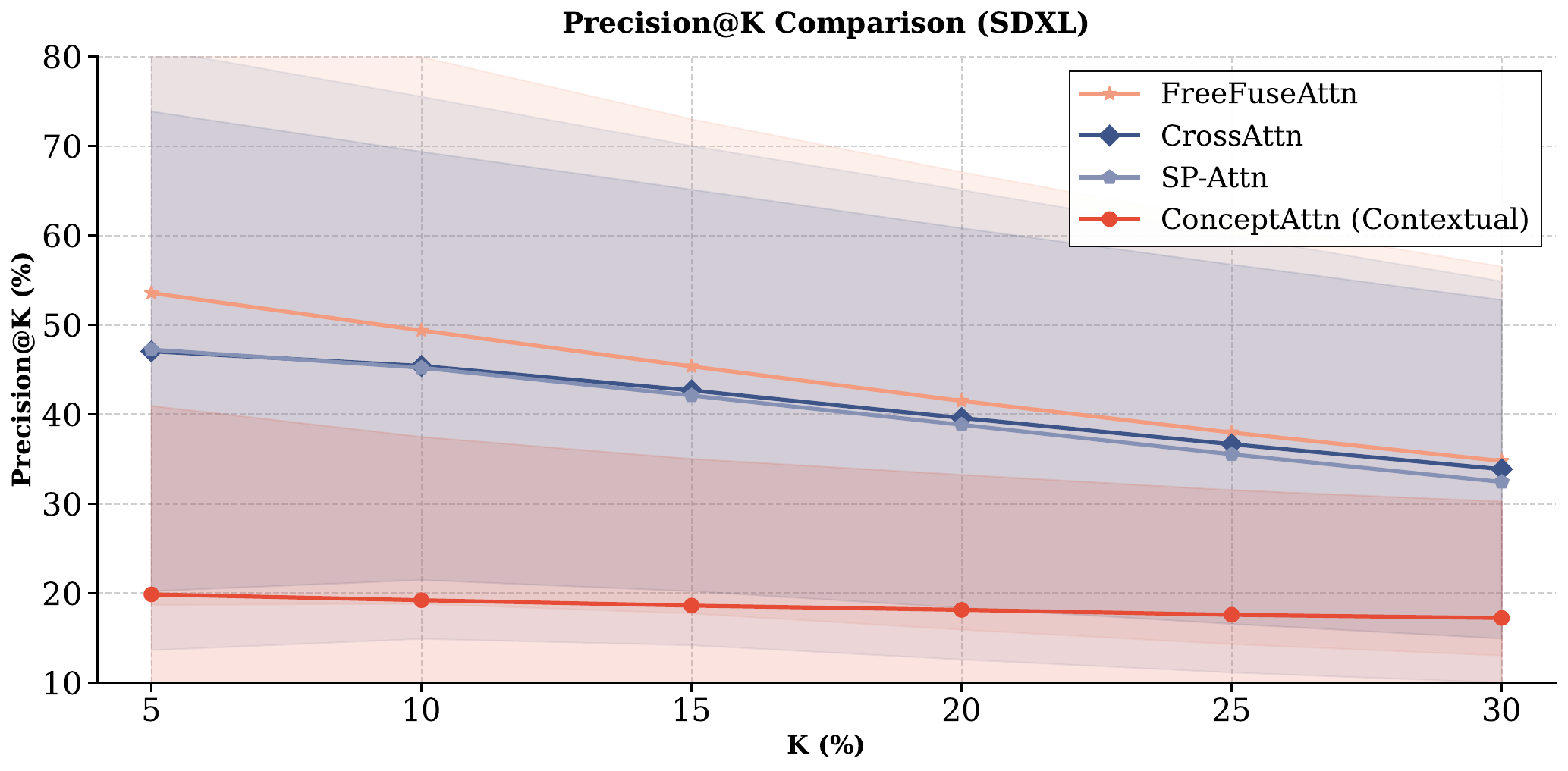}
  \captionsetup{width=1\linewidth}
  \caption{On SDXL, we evaluate the spatial alignment of different mechanisms against SAM3-generated ground truth masks over 300 samples. \textbf{Precision@K} measures the percentage of the top $K\%$ activated tokens that correctly fall within the subject's region. FreeFuseAttn demonstrates superior localization accuracy.}
  \label{fig:precision_curve_sdxl}
\end{figure}

These choices were determined through the same experimental procedures as those described in the main text; please refer to Fig.~\ref{fig:sdxl_choices} and Fig.~\ref{fig:precision_curve_sdxl}.

\subsection{Qualitative Analysis}

\begin{figure}[t]
  \centering
  \centering
  \includegraphics[width=1\textwidth]{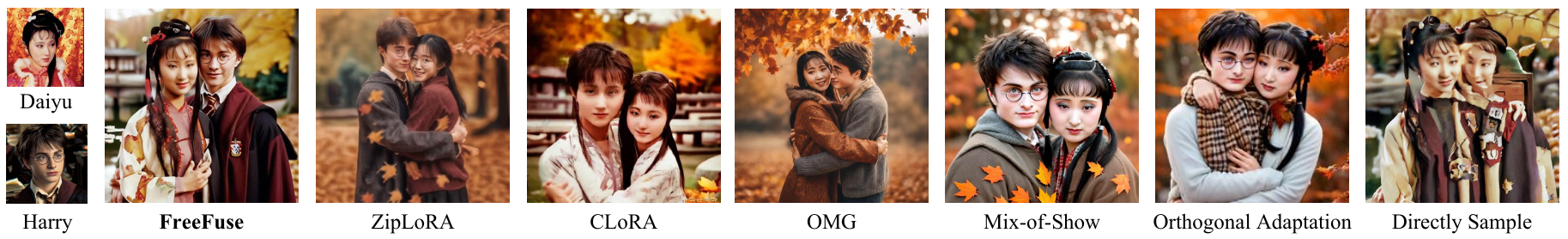}
  \captionsetup{width=1\linewidth}
  \caption{Qualitative results of FreeFuse applied to SDXL. Prompt: \textit{harry\_potter and daiyu\_lin, both faces close together, autumn leaves blurred in the background}}
  \label{fig:sdxl_comp}
\end{figure}

\begin{figure}[t]
  \centering
  \centering
  \includegraphics[width=1\textwidth]{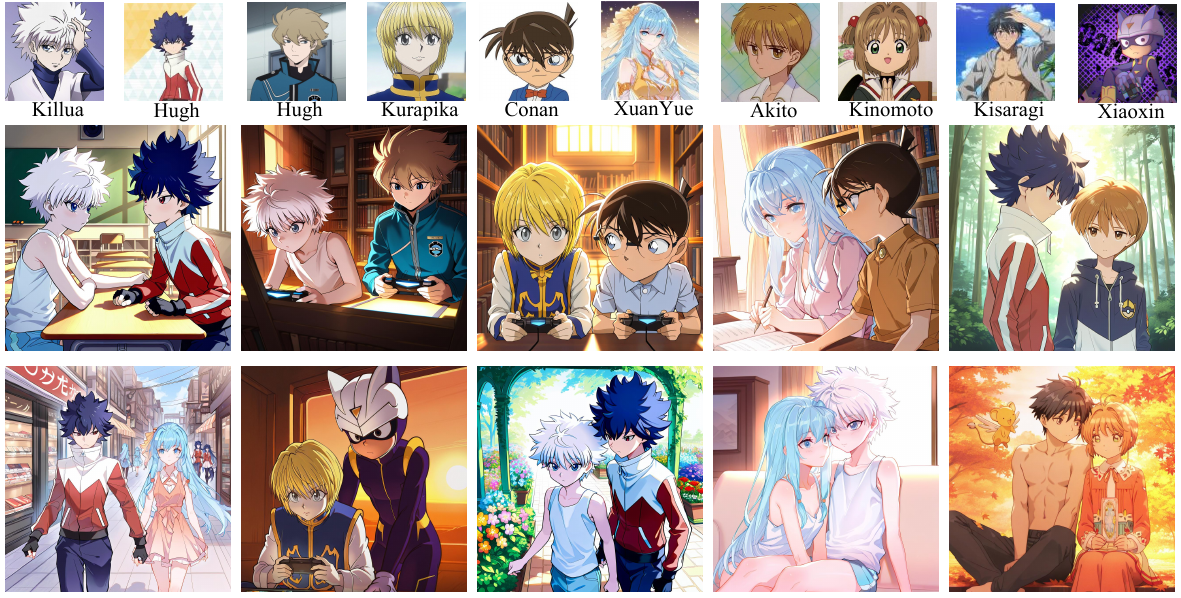}
  \captionsetup{width=1\linewidth}
  \caption{The results obtained from the SDXL-derived Illustrious-XL model further demonstrate the generalizability of our method.}
  \label{fig:sdxl_more_res}
\end{figure}

Fig.~\ref{fig:sdxl_comp} presents the qualitative results of FreeFuse applied to SDXL. The results demonstrate that our method performs well on SDXL, generating high-quality images without requiring any control signals other than text prompts. We further present the generation results on Illustrious-XL, currently the most popular SDXL-derived model on Civitai in Fig.~\ref{fig:sdxl_more_res}.

\noindent\textbf{Note on Performance:} \textit{While FreeFuse effectively mitigates concept bleeding in SDXL, we observe that the overall prompt adherence and spatial complexity handling of the base SDXL model are naturally lower than that of Flux. Consequently, our main quantitative benchmarks in Section 4 focus on the Flux architecture to decouple the fusion algorithm's performance from the base model's capabilities.}

\section{Post-processing Details}
\label{sec:appendix_postprocessing}

To translate the similarity maps into coherent and exclusive binary masks, we design a two-stage post-processing pipeline. This pipeline ensures topological integrity for the foreground regions and enforces strict semantic segregation between competing subjects.

\subsection{Morphological Refinement}
\label{subsec:morphological}

\begin{figure}[t]
  \centering
  \centering
  \includegraphics[width=1\textwidth]{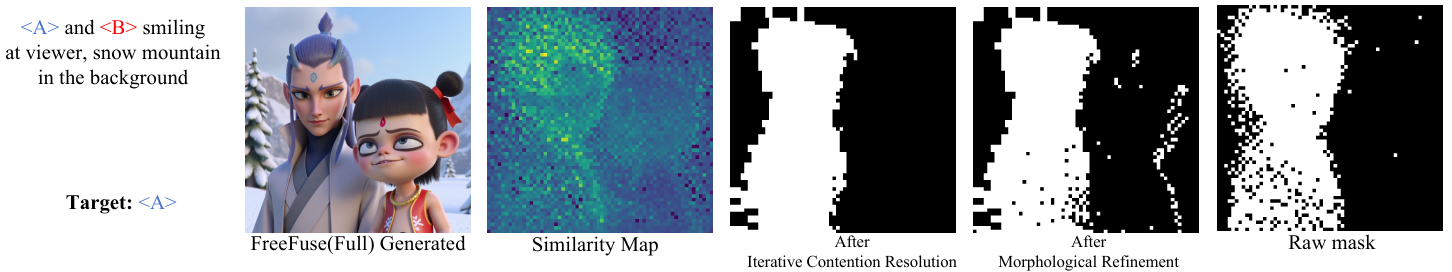}
  \captionsetup{width=1\linewidth}
  \caption{Post-processing plays a crucial role in further eliminating hollow artifacts within the masks and enhancing the overall stability of the method.}
  \label{fig:postprocess_importance}
\end{figure}

\begin{table}[t]
\centering
\caption{\rev{Hyperparameter settings for FreeFuse mask extraction and post-processing.}}
\label{tab:hyperparams}
\begin{tabular}{lcc}
\toprule
\textbf{Parameter} & \textbf{Symbol} & \textbf{Value} \\
\midrule
\rowcolor{revblue} Mask Extraction Steps & $K$ & $5$ \\
\rowcolor{revblue} Inference Denoising Steps & $T_{\mathrm{denoise}}$ & $28$ \\
\rowcolor{revblue} FreeFuseAttn Block & $b$ & $18$ \\
\rowcolor{revblue} FreeFuseAttn Temperature & $\tau$ & $4000$ \\
\rowcolor{revblue} TopK Anchor Ratio & $\rho$ & $10\%$ \\
\rowcolor{revblue} TopK Anchors & $k_{\mathrm{anchor}}$ & $\max(1,\lfloor \rho N \rfloor)$ \\
\rowcolor{revblue} Morphological Kernel & $\mathcal{K}$ & $\begin{smallmatrix}1&1\\1&1\end{smallmatrix}$ \\
\rowcolor{revblue} Router Iterations & $T_{\mathrm{route}}$ & $15$ \\
Momentum & $\mu$ & $0.2$ \\
Gravity Weight & $\lambda_g$ & $2 \times 10^{-5}$ \\
Spatial Voting Weight & $\lambda_s$ & $2 \times 10^{-5}$ \\
\bottomrule
\end{tabular}
\end{table}

To distinguish the foreground from the background, we apply a sequence of morphological operations to the binary foreground mask $M_{raw}$.

\rev{Let $\mathcal{K}$ denote the $2\times2$ structural element}
\begin{equation}
\colorbox{revblue}{$\displaystyle
    \mathcal{K} =
    \begin{bmatrix}
    1 & 1 \\
    1 & 1
    \end{bmatrix}.
$}
\end{equation}
\rev{We define the \textit{Opening} operation ($\circ$) to remove isolated noise pixels, followed by the \textit{Closing} operation ($\bullet$) to fill interior holes:}

\begin{equation}
    M_{opened} = M_{raw} \circ \mathcal{K} = (M_{raw} \ominus \mathcal{K}) \oplus \mathcal{K}
\end{equation}
\begin{equation}
    M_{clean} = M_{opened} \bullet \mathcal{K} = (M_{opened} \oplus \mathcal{K}) \ominus \mathcal{K}
\end{equation}

\rev{where $\oplus$ and $\ominus$ denote Dilation and Erosion, respectively. This process yields a topologically clean foreground mask $M_{clean}$ that delineates the union of all subjects against the background.}

\subsection{Iterative Contention Resolution (Router)}
\label{subsec:router_algorithm}

Within the foreground region, multiple subject LoRAs may compete for the same spatial tokens. To resolve these conflicts and assign each token $p$ to a unique subject $c \in \{1, \dots, C\}$, we propose an Iterative Routing algorithm. Unlike standard $\text{argmax}$ or $\text{Softmax}$, which suffer from "winner-takes-all" instability and lack spatial awareness, our method incorporates spatial cohesion constraints and adaptive load balancing.

We initialize the routing logits $L^{(0)}_{p,c}$ with the similarity scores derived from FreeFuseAttn. The algorithm updates these logits iteratively over $T$ steps (default $T=15$). At each step $t$, the update rule is composed of four terms:

\paragraph{1. Linear Normalization.} To prevent exponential suppression of weaker signals (a common issue with Softmax), we normalize logits linearly to the range $[0, 1]$:
\begin{equation}
    P_{p,c}^{(t)} = \frac{L_{p,c}^{(t)} - \min_c(L_{p,c}^{(t)})}{\max_c(L_{p,c}^{(t)}) - \min_c(L_{p,c}^{(t)}) + \epsilon}
\end{equation}

\paragraph{\rev{2. Momentum Averaging.}}
\rev{We maintain a running average of these probabilities $\bar{P}^{(t)}$ with momentum $\mu=0.2$ to stabilize the trajectory.}

\paragraph{3. Spatial Cohesion (Gravity).} We calculate the dynamic centroid $(\bar{x}_c, \bar{y}_c)$ for each subject based on the soft distribution $\bar{P}^{(t)}$. A penalty is applied proportional to the squared Euclidean distance from the centroid to enforce compactness:
\begin{equation}
    \mathcal{E}_{gravity}(p, c) = \lambda_{g} \cdot \| \mathbf{u}_p - (\bar{x}_c, \bar{y}_c) \|^2
\end{equation}
where $\mathbf{u}_p$ is the spatial coordinate of token $p$ ranging from $-1$ to $1$.

\paragraph{4. Local Neighborhood Voting.} To encourage local smoothness, we aggregate votes from the $3 \times 3$ neighborhood $\mathcal{N}(p)$:
\begin{equation}
    \mathcal{V}_{spatial}(p, c) = \lambda_{s} \sum_{q \in \mathcal{N}(p)} \bar{P}_{q,c}^{(t)}
\end{equation}

\paragraph{Final Update Rule.} Combining these components, the logits are updated as:
\begin{equation}
    L_{p,c}^{(t+1)} = L_{p,c}^{(t)} + \mathcal{V}_{spatial}(p, c) - \mathcal{E}_{gravity}(p, c)
\end{equation}
After $T$ iterations, the final subject mask is obtained via $\text{argmax}_c L_{p,c}^{(T)}$. This ensures that the generated masks are not only semantically grounded but also spatially compact.

For reproducibility, we list the specific hyperparameters used in our experiments in Tab.~\ref{tab:hyperparams}.

Fig.~\ref{fig:postprocess_importance} provides an intuitive example, demonstrating that our proposed post-processing plays a crucial role in enhancing the cohesion and stability of the masks.

\section{Prompts and LoRAs for Experiments}
\label{sec:prompts}

To ensure the reproducibility of our experiments, we provide the full list of scene prompts used in the multi-subject interaction stress tests. These prompts serve as the environmental context into which the specific subject tokens (e.g., \texttt{<Subject A>}, \texttt{<Subject B>}, etc.) are integrated.

\begin{tcolorbox}[colback=gray!5!white, colframe=gray!50!black, title=Scene Prompts for 4-Subject Generation]
\footnotesize
\begin{multicols}{2}
\begin{enumerate}[itemsep=2pt, topsep=2pt, leftmargin=*]
\tiny

\item \texttt{Four characters talking on a riverside promenade lined with cherry trees, petals falling into the water, soft morning sunlight illuminating their smiles.}
\item \texttt{Four friends sitting on a rustic porch of a countryside cottage, chatting animatedly, dappled sunlight filtering through the hanging vines.}
\item \texttt{Four characters standing on a vast frozen lake in the wilderness, bundled in heavy coats, steam rising from their breath, golden hour lighting reflecting on the ice.}
\item \texttt{Four people having a serious conversation on a tree-lined city boulevard, surrounded by fallen orange and red leaves, overcast sky creating a moody atmosphere.}
\item \texttt{Four characters talking on a busy downtown street corner in the rain, huddled under umbrellas, wet asphalt reflecting neon signs, cinematic cool tones.}
\item \texttt{Four teenagers hanging out on a rooftop terrace overlooking the city, vibrant sunset colors painting the sky, nostalgic vibe.}
\item \texttt{Four characters walking down an ancient cobblestone alley, engaged in deep discussion, long shadows stretching out behind them in the late afternoon sun.}
\item \texttt{Four figures whispering on a mist-covered harbor dock at dawn, mysterious atmosphere, soft focus background of docked ships.}
\item \texttt{Four professionals in suits discussing business on a high-rise office balcony, modern city skyline in the background, bright clear blue sky.}
\item \texttt{Four characters having a picnic in a sunny wildflower meadow near a cliff edge, talking and laughing, surrounded by colorful blooming nature.}
\item \texttt{Four people arguing heatedly on a desolate windswept beach, hair blowing in the wind, storm clouds gathering above the ocean, dramatic high-contrast lighting.}
\item \texttt{Four characters talking by a stone fountain in an old European town square, vintage atmosphere, warm sunlight glowing on the water.}
\item \texttt{Four joggers taking a break on a scenic coastal road, dewdrops on the roadside grass, fresh and energetic atmosphere.}
\item \texttt{Four characters sitting in a circle on the sand of a quiet beach at night, illuminated by the glow of a nearby campfire, cozy and intimate setting.}
\item \texttt{Four silhouettes talking on a pedestrian bridge over a city canal at twilight, reflection visible in the water below, purple and blue gradient sky.}
\item \texttt{Four students studying and talking in a university library courtyard, seated under a large oak tree, scattered books, warm afternoon light.}
\item \texttt{Four characters talking while walking their dogs along a quiet suburban sidewalk, dynamic poses, bright and cheerful daylight.}
\item \texttt{Four elderly characters sitting at a table outside a street corner cafe, talking reminiscences, peaceful atmosphere, soft golden lighting.}
\item \texttt{Four characters chatting in a neon-lit back alley of a cyberpunk city, glowing holographic ads around them, rain-slicked surfaces.}
\item \texttt{Four curious characters looking at a map in a dense jungle clearing, sunbeams breaking through the thick canopy, adventure theme.}

\end{enumerate}
\end{multicols}
\end{tcolorbox}

\begin{tcolorbox}[colback=gray!5!white, colframe=gray!50!black, title=Character-Object Interaction Prompts]
\footnotesize
\textit{Note: In these prompts, \texttt{<Subject>} is replaced by the specific character token.}

\begin{multicols}{2}
\begin{enumerate}[itemsep=4pt, topsep=2pt, leftmargin=*]
\tiny

\item \textbf{Tactical Pistol:} \texttt{<Subject> aiming A modern semi-automatic pistol with a threaded suppressor attachment, featuring a two-tone black and silver finish.}

\item \textbf{Golden Snitch:} \texttt{<Subject> reaching out hand to catch A gleaming golden snitch with intricately ribbed wings spread wide.}

\item \textbf{Inception Top:} \texttt{<Subject> staring intently at A sleek black spinning inception top balanced on its needle-sharp point spinning flawlessly on the table.}

\item \textbf{Nike SB Sneaker:} \texttt{<Subject> sitting on the ground and holding a Nike SB sneaker in classic black, white, and red colorway.}

\item \textbf{Armored Battle Car:} \texttt{<Subject> standing casually in front of A heavily armored tactical vehicle with angular matte gray plating.}

\end{enumerate}
\end{multicols}
\end{tcolorbox}

\begin{figure}[t]
  \centering
  \centering
  \includegraphics[width=0.9\textwidth]{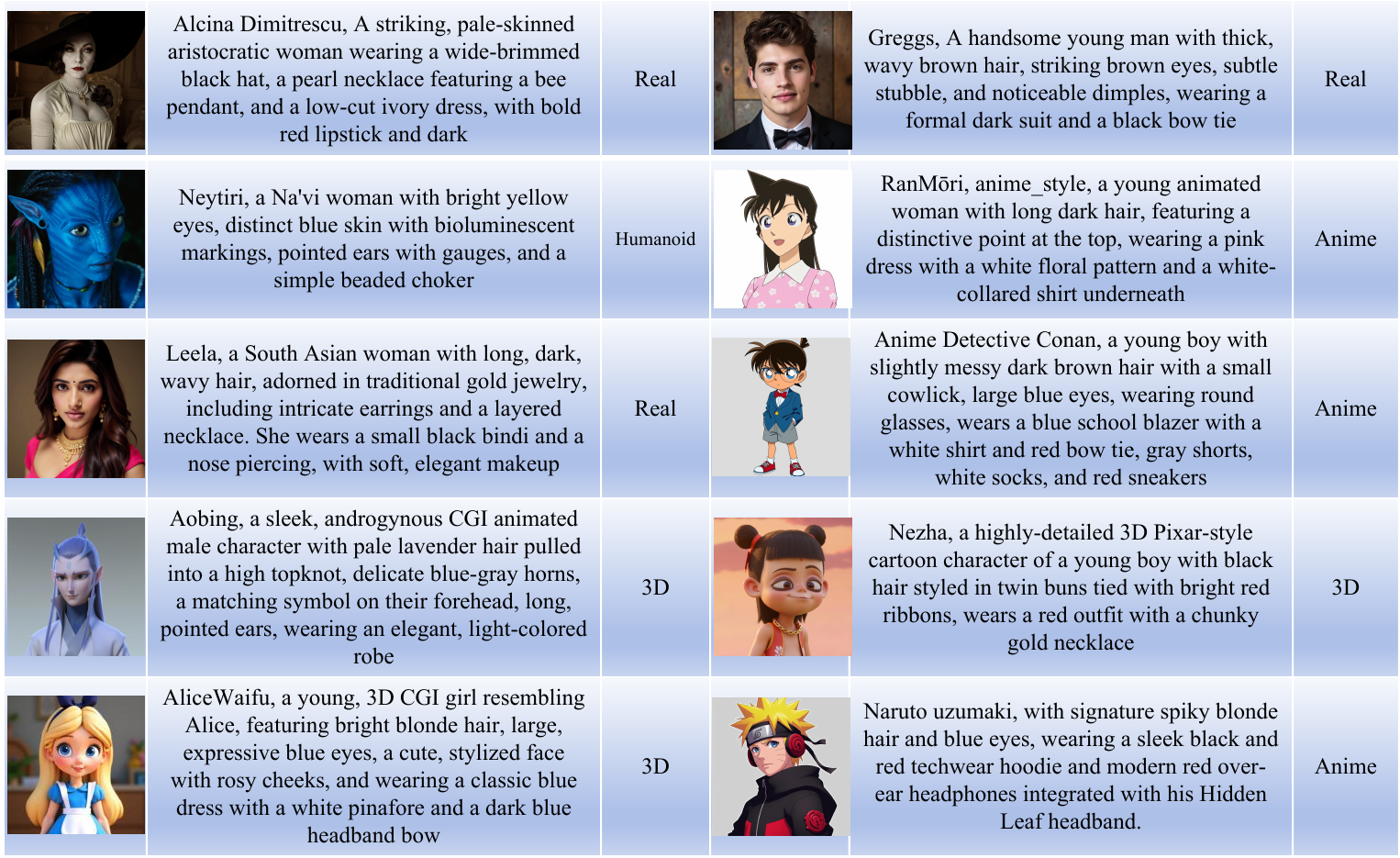}
  \captionsetup{width=1\linewidth}
  \caption{Character LoRAs used in quantitative and qualitative comparisons.}
  \label{fig:our_loras}
\end{figure}

\begin{figure}[t]
  \centering
  \centering
  \includegraphics[width=0.9\textwidth]{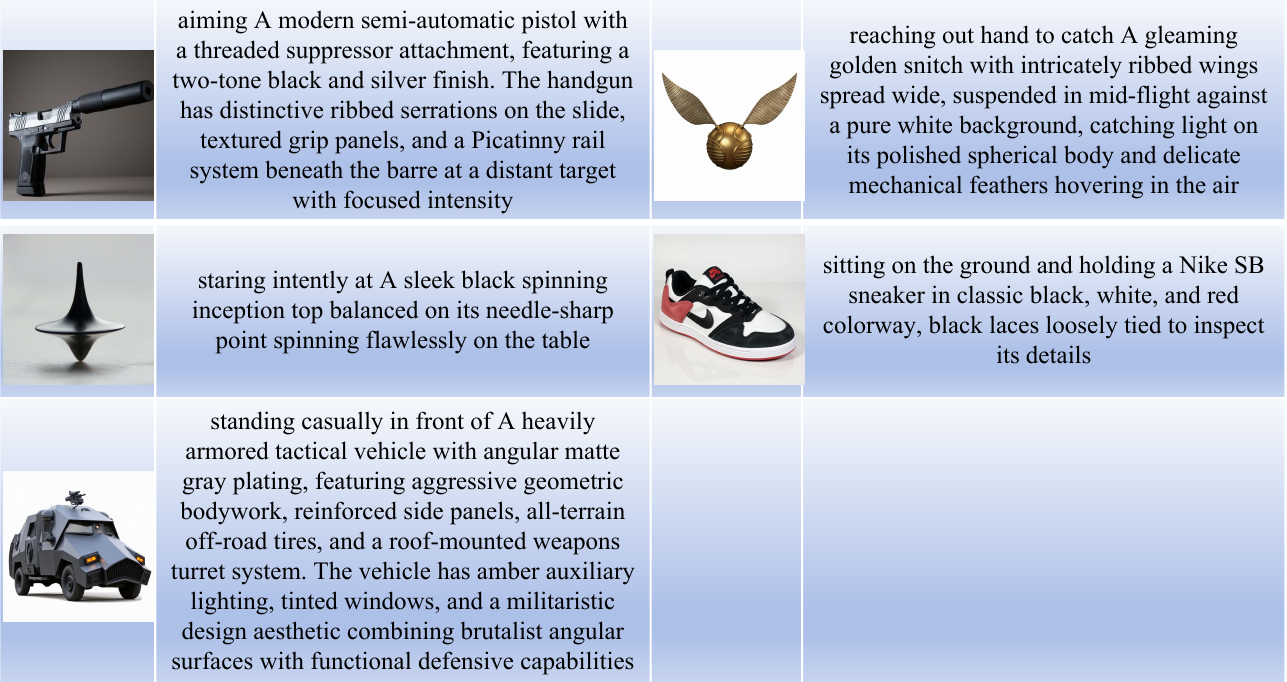}
  \captionsetup{width=1\linewidth}
  \caption{Object LoRAs used in quantitative and qualitative comparisons.}
  \label{fig:our_loras_obj}
\end{figure}

To facilitate reproducibility and ensure transparency, we list the 15 LoRA checkpoints utilized in our quantitative and qualitative comparisons. The dataset consists of 10 Character LoRAs and 5 Object LoRAs, selected to cover a diverse range of styles (photorealistic, anime, 3D) and structural complexities (rigid bodies, deformable objects), as illustrated in Fig.~\ref{fig:our_loras} and Fig.~\ref{fig:our_loras_obj}.

All checkpoints were obtained from open-source repositories (e.g., Civitai, Hugging Face). Upon the publication of this paper, we will release the specific weight files and the corresponding retrieval scripts to the community.

\section{More Qualitative Comparisons}
\label{sec:more_comp}

\begin{figure}[t]
  \centering
  \centering
  \includegraphics[width=1\textwidth]{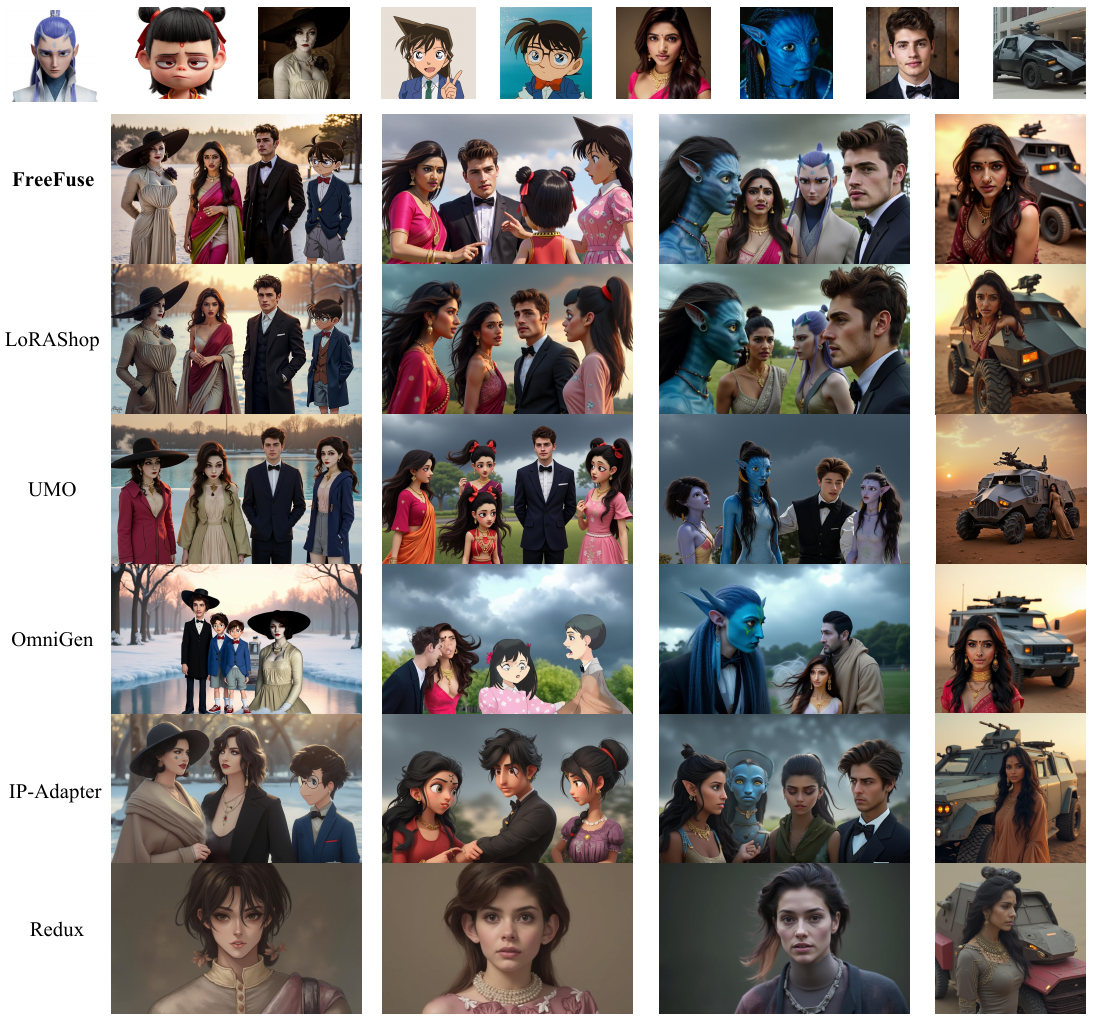}
  \captionsetup{width=1\linewidth}
  \caption{Additional Qualitative Comparisons.}
  \label{fig:more_comp}
\end{figure}

We further provide more qualitative comparisons in Fig.~\ref{fig:more_comp}.

\section{\texorpdfstring{\rev{Scaling to More LoRAs}}{Scaling to More LoRAs}}
\label{sec:lora_count_scaling}

\rev{To evaluate whether FreeFuse remains effective as the number of simultaneously loaded LoRAs increases, we construct a controlled LoRA-count ablation with $k\in\{2,3,4,5\}$ subject LoRAs. The baseline uses the same LoRAs and prompts with naive simultaneous LoRA activation, while FreeFuse uses the proposed mask extraction, routing, and attention bias. As shown in Tab.~\ref{tab:lora_count_scaling}, identity preservation becomes increasingly difficult as more adapters compete, but FreeFuse consistently improves ArcFace ID over the baseline at every LoRA count, with an average gain of $+0.3207$. HPSv2 remains comparable or slightly higher, indicating that the identity gain does not come from degrading global image preference.}

\begin{table*}[t]
\centering
\small
\caption{\rev{Quality trends as the number of simultaneously loaded LoRAs increases. Baseline denotes direct multi-LoRA activation without FreeFuse routing.}}
\label{tab:lora_count_scaling}
\resizebox{0.86\textwidth}{!}{%
\begin{tabular}{c|ccc|ccc}
\toprule
\multirow{2}{*}{\textbf{\# LoRAs}} &
\multicolumn{3}{c|}{\textbf{ArcFace ID} $\uparrow$} &
\multicolumn{3}{c}{\textbf{HPSv2} $\uparrow$} \\
& Baseline & FreeFuse & $\Delta$ & Baseline & FreeFuse & $\Delta$ \\
\midrule
\rowcolor{revblue} 2 & 0.4305 & \textbf{0.7002} & +0.2697 & 0.2806 & \textbf{0.2806} & +0.0000 \\
\rowcolor{revblue} 3 & 0.2695 & \textbf{0.6623} & +0.3928 & 0.2786 & \textbf{0.2804} & +0.0019 \\
\rowcolor{revblue} 4 & 0.2028 & \textbf{0.6285} & +0.4257 & 0.2776 & \textbf{0.2809} & +0.0033 \\
\rowcolor{revblue} 5 & 0.1759 & \textbf{0.3705} & +0.1946 & 0.2712 & \textbf{0.2773} & +0.0061 \\
\midrule
\rowcolor{revblue} Avg. & 0.2697 & \textbf{0.5904} & +0.3207 & 0.2770 & \textbf{0.2798} & +0.0028 \\
\bottomrule
\end{tabular}}
\end{table*}

\begin{figure*}[t]
  \centering
  \begin{subfigure}{0.48\textwidth}
    \centering
    \includegraphics[width=\linewidth]{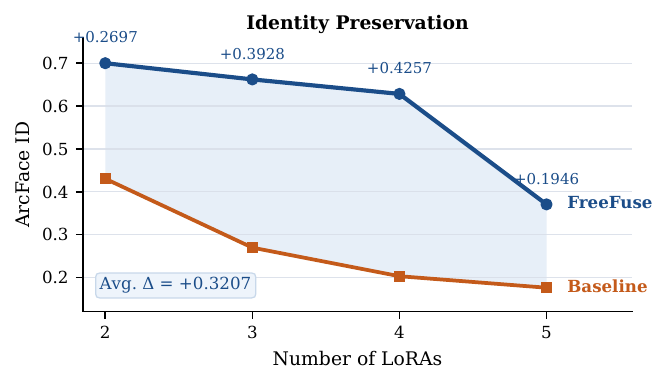}
    \caption{\rev{ArcFace ID}}
  \end{subfigure}
  \hfill
  \begin{subfigure}{0.48\textwidth}
    \centering
    \includegraphics[width=\linewidth]{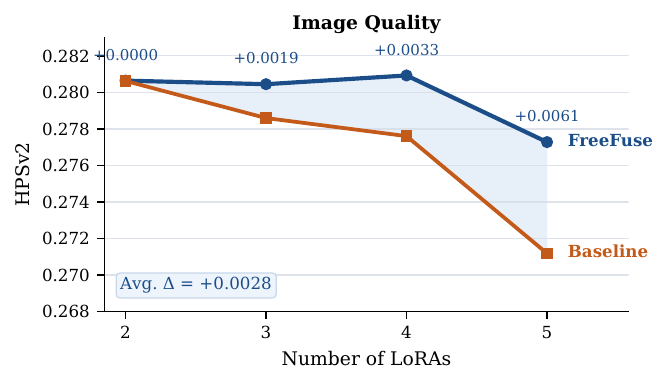}
    \caption{\rev{HPSv2}}
  \end{subfigure}
  \captionsetup{width=0.95\linewidth}
  \caption{\rev{Trends under increasing LoRA count. FreeFuse preserves identity substantially better than direct multi-LoRA activation, while maintaining comparable global preference scores.}}
  \label{fig:lora_count_scaling}
\end{figure*}

\FloatBarrier

\section{\texorpdfstring{\rev{Additional Evaluation on a Public Dual-Subject Protocol}}{Additional Evaluation on a Public Dual-Subject Protocol}}
\label{sec:lorashop_protocol}

\rev{We further evaluate FreeFuse on the public dual-subject protocol of LoRAShop~\citep{dalva2025lorashop}, whose LoRA weights and prompting setup are directly reproducible. Tab.~\ref{tab:lorashop_protocol} reports the comparison under this protocol. FreeFuse achieves the best identity preservation and prompt alignment, while remaining competitive in human-preference and aesthetic metrics. This protocol is complementary to our main benchmark: LoRAShop primarily evaluates two real-person LoRAs, whereas our main benchmark includes photorealistic, anime, 3D, object, garment, and accessory LoRAs, and stresses up to four simultaneous LoRAs in a single prompt.}

\begin{table}[t]
\centering
\small
\caption{\rev{Evaluation on the public LoRAShop dual-subject protocol.}}
\label{tab:lorashop_protocol}
\resizebox{\linewidth}{!}{%
\begin{tabular}{lcccc}
\toprule
\textbf{Method} & \textbf{ArcFace ID} $\uparrow$ & \textbf{CLIP-T bigG} $\uparrow$ & \textbf{HPSv2} $\uparrow$ & \textbf{Aesthetic v2} $\uparrow$ \\
\midrule
\rowcolor{revblue}
\textbf{FreeFuse} & \textbf{0.6557} & \textbf{0.5184} & 0.3117 & 5.7426 \\
\rowcolor{revblue} IP-Adapter & 0.2139 & 0.4826 & 0.3100 & \textbf{6.2833} \\
\rowcolor{revblue} OmniGen & 0.5554 & 0.5032 & 0.3040 & 6.0466 \\
\rowcolor{revblue} LoRAShop & 0.6376 & 0.5157 & \textbf{0.3153} & 5.9921 \\
\rowcolor{revblue} FLUX Redux & 0.2478 & 0.2508 & 0.2527 & 6.1375 \\
\rowcolor{revblue} UMO & 0.5091 & 0.4991 & 0.3070 & 6.0185 \\
\bottomrule
\end{tabular}}
\end{table}

\FloatBarrier

\section{User Study Details}
\label{sec:user_study}

To complement our quantitative metrics, we conducted a blind user preference study to evaluate the perceptual quality of the generated images. The study focused on two core capabilities: multi-subject interaction (Q1) and character-object interaction (Q2). It includes 17 participants, aged 18-40, spanning undergraduate students, master's students, PhD students, and professors. The questionnaire uses blind presentation for all methods, and the study contains 40 cases in total (20 character-character and 20 character-object cases). The final score is obtained by averaging participant rankings across cases.

Participants were presented with anonymized outputs from FreeFuse and baseline methods (shuffled to prevent bias) and were asked to perform a ranking task. The questionnaire instructions and evaluation criteria were designed as follows:

\begin{tcolorbox}[colback=gray!5!white, colframe=gray!50!black, title=Questionnaire Instruction]
\small
\textbf{Task:} Please rank the set of generated images displayed below. The reference character images are provided as \texttt{<A>}, \texttt{<B>}, \texttt{<C>}, and \texttt{<D>}.

\vspace{2mm}
\textbf{Ranking Criteria (in descending order of priority):}
\begin{enumerate}[leftmargin=*]
    \item \textbf{Identity Preservation:} How accurately does the generated character resemble the target reference image? (Higher consistency = Higher Rank)
    \item \textbf{Text-Image Alignment:} Does the image accurately reflect the content described in the prompt? (Higher consistency = Higher Rank)
    \item \textbf{Aesthetic Quality:} How visually appealing and natural is the overall image? (Higher quality = Higher Rank)
\end{enumerate}

\vspace{2mm}
\textbf{Example Prompt (Q1 - Multi-Character):}
\textit{``<A>, <B>, <C>, and <D> are having a friendly conversation by the fountain pool.''}
\end{tcolorbox}

For the Character-Object interaction (Q2), the same criteria were applied, with "Identity Preservation" extending to the structural accuracy of the specific object LoRA. The aggregated rankings were converted into the preference scores reported in Tab.~\ref{tab:compare}.

\section{Latency and Computational Cost Analysis}
\label{sec:latency_analysis}

To evaluate the practical efficiency of FreeFuse, we measured the inference latency across varying numbers of subject LoRAs (from 2 to 6). All experiments were conducted on a single L20 (48GB). The baseline represents the standard FLUX.1-dev inference with naive LoRA sampling.

\textbf{Experimental Protocol.} We define the total inference time as the duration from the initial prompt encoding to the final pixel decoding. For FreeFuse, this includes the additional Phase 1 (Mask Extraction) and the Iterative Contention Resolution.

\begin{table}[h]
\centering
\caption{Inference latency comparison (in seconds) at $1024\times1024$ resolution. \textbf{Comparison Key:} While Naive Flux is faster, it fails to generate coherent images. Compared to the leading training-free competitor LoRAShop~\cite{dalva2025lorashop}, FreeFuse achieves a $1.8\times \sim 2.1\times$ speedup, offering a superior trade-off between fidelity and efficiency.}
\label{tab:latency}
\vspace{2mm}
\resizebox{1.0\textwidth}{!}{%
\begin{tabular}{@{}lccccc@{}}
\toprule
\textbf{Method} & \textbf{2 LoRAs} & \textbf{3 LoRAs} & \textbf{4 LoRAs} & \textbf{5 LoRAs} & \textbf{6 LoRAs} \\ \midrule
\textcolor{gray}{Naive Flux (Fails on complex scenes)} & \textcolor{gray}{21.95s} & \textcolor{gray}{24.60s} & \textcolor{gray}{27.11s} & \textcolor{gray}{29.64s} & \textcolor{gray}{32.08s} \\ \midrule
LoRAShop \cite{dalva2025lorashop} & 64.78s & 84.36s & 103.60s & 122.96s & 142.27s \\
\textbf{FreeFuse (Ours)} & \textbf{36.08s} & \textbf{42.96s} & \textbf{50.29s} & \textbf{58.63s} & \textbf{67.96s} \\ \midrule
\rowcolor{green!10} \textbf{Speedup vs. LoRAShop} & \textbf{+44.3\%} & \textbf{+49.1\%} & \textbf{+51.5\%} & \textbf{+52.3\%} & \textbf{+52.2\%} \\ 
\bottomrule
\end{tabular}%
}
\end{table}

\noindent\textbf{Note on Efficiency Trade-offs:} 
While the naive Flux baseline exhibits lower latency, it fails to produce semantically valid outputs in multi-subject scenarios due to severe feature conflict (as evidenced in Fig.~\ref{fig:teaser}). Consequently, the computational overhead of FreeFuse is a necessary trade-off for ensuring identity preservation. Furthermore, when benchmarked against LoRAShop~\cite{dalva2025lorashop}, the leading training-free alternative, FreeFuse demonstrates significantly better scalability. As the number of subjects increases to six, LoRAShop's inference time rises linearly to over 140 seconds, whereas FreeFuse remains under 70 seconds. This indicates that our framework provides a more practical balance between high-fidelity generation and runtime efficiency.

\end{document}